\documentclass[10pt,twocolumn,letterpaper]{article}

\usepackage[pagenumbers]{cvpr} 

\usepackage{soul}
\usepackage{epigraph}
\usepackage{amsmath}
\usepackage{bm}

\usepackage{colortbl}       
\usepackage{xcolor}         

\usepackage{multirow, tabularx}
\usepackage{makecell}
\usepackage{booktabs} 
\usepackage{arydshln} 
\usepackage{array} 
\usepackage{tcolorbox}

\usepackage{diagbox}
\usepackage{bbding} 
\usepackage{subfloat}
\usepackage{xspace} 

\def\eg{\emph{e.g.}}
\def\ie{\emph{i.e.}}

\newcommand{\mysubsubsec}[1]{\noindent\textbf{#1}}
\newcommand{\piref}{\pi_\text{ref}}

\definecolor{gray_bg}{gray}{0.9} 

\newcommand\correspondingauthor{\thanks{Corresponding author. }}

\definecolor{cvprblue}{rgb}{0.21,0.49,0.74}
\usepackage[pagebackref,breaklinks,colorlinks,allcolors=cvprblue]{hyperref}

\makeatletter
\newcommand{\thickhline}{
	\noalign {\ifnum 0=`}\fi \hrule height 1pt
	\futurelet \reserved@a \@xhline
}
\makeatother

\def\paperID{255} 
\def\confName{CVPR}
\def\confYear{2025}

\title{SILMM: Self-Improving Large Multimodal Models for Compositional Text-to-Image Generation}

\author{Leigang Qu$^1$, Haochuan Li$^{1}$, Wenjie Wang$^2$\correspondingauthor, Xiang Liu$^1$, Juncheng Li$^{3}$, Liqiang Nie$^4$, Tat-Seng Chua$^1$\\
\normalsize$^1$National University of Singapore, 
$^2$University of Science and Technology of China, 
$^3$Zhejiang University, \\
\normalsize$^4$Harbin Institute of Technology (Shenzhen)\\
{\tt\small leigangqu@gmail.com, haochuan@u.nus.edu, wenjiewang96@gmail.com, liu.xiang@u.nus.edu} \\
{\tt\small junchengli@zju.edu.cn, nieliqiang@gmail.com, dcscts@nus.edu.sg}
}

\begin{document}
\maketitle

\begin{abstract} 
Large Multimodal Models (LMMs) have demonstrated impressive capabilities in multimodal understanding and generation, pushing forward advancements in text-to-image generation.
However, achieving accurate text-image alignment for LMMs, particularly in compositional scenarios, remains challenging. 
Existing approaches, such as layout planning for multi-step generation and learning3 from human feedback or AI feedback, depend heavily on prompt engineering, costly human annotations, and continual upgrading, limiting flexibility and scalability. 
In this work, we introduce a model-agnostic iterative self-improvement framework (\textbf{SILMM}) that can enable LMMs to provide helpful and scalable self-feedback and optimize text-image alignment via Direct Preference Optimization (DPO). 
DPO can readily applied to LMMs that use discrete visual tokens as intermediate image representations; while it is less suitable for LMMs with continuous visual features, as obtaining generation probabilities is challenging.
To adapt SILMM to LMMs with continuous features, we propose a diversity mechanism to obtain diverse representations and a kernel-based continuous DPO for alignment. 
Extensive experiments on three compositional text-to-image generation benchmarks validate the effectiveness and superiority of SILMM, showing improvements exceeding 30\% on T2I-CompBench++ and around 20\% on DPG-Bench. The code is available at \url{https://silmm.github.io/}.

\end{abstract}    
\section{Introduction}
\label{sec:intro}
\begin{figure}[t]
\setlength{\abovecaptionskip}{0.05cm}
\setlength{\belowcaptionskip}{0cm}
\centering
\includegraphics[width=0.48\textwidth]{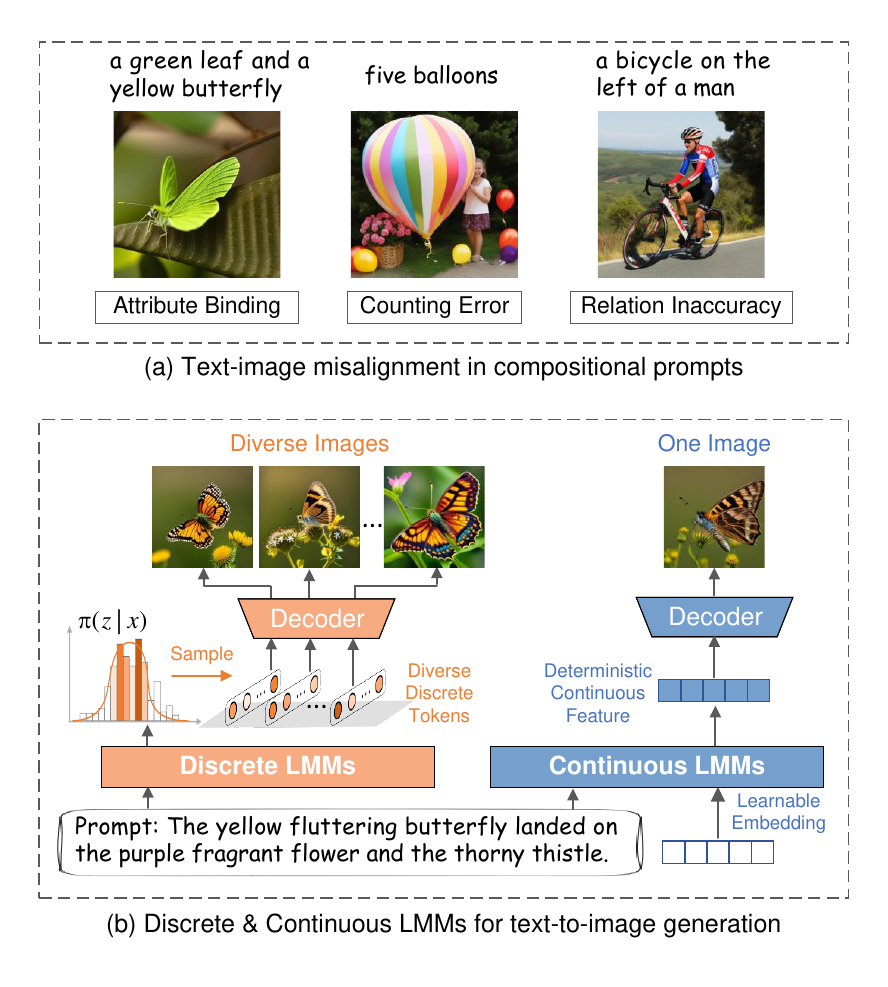}
\vspace{-5ex}
\caption{Illustration of (a) text-image misalignment in compositional prompts and (b) comparison of discrete and continuous LMMs for T2I. Given a prompt, discrete LMMs can sample diverse token sequences from categorical distributions, while continuous LMMs can only produce a single deterministic feature vector. Note that the input learnable embeddings are optional for some continuous LMMs~\cite{sun2024generative}. }
\label{fig:intro}
\vspace{-3ex}
\end{figure}

Large Multimodal Models (LMMs) are advancing rapidly, surpassing Large Language Models (LLMs) by embracing multimodal capabilities for multimodal content perception, understanding~\cite{qu2021dynamic, li2022invariant, li2023transformer}, and generation~\cite{sun2024generative, ge2024making}.  
In particular, LMMs demonstrate promising abilities in interpreting user input prompts for text-to-image generation (T2I)~\cite{ramesh2022hierarchical, saharia2022photorealistic}, producing vivid and photorealistic images. 
However, as shown in Fig.~\ref{fig:intro}(a), achieving precise \textit{text-image alignment} between generated images and complex prompts remains challenging, especially for compositional prompts involving multiple objects, attributes, counting, and complex relationships~\cite{qu2023layoutllm, feng2023training, chefer2023attend}.

To enhance text-image alignment, existing work falls into two primary research lines. One line focuses on decomposing the T2I task into multiple stages. For example, some methods perform layout planning before generating the image~\cite{feng2023layoutgpt, lian2023llm, yang2024mastering}; while some split the image into sections for multi-step generation via multi-agent collaboration~\cite{wang2024divide, qin2024diffusiongpt}. 
However, these methods depend on extensive multi-step prompt engineering, which risks error accumulation. 
The second research line emphasizes learning from human feedback (RLHF~\cite{ouyang2022training}) to improve text-image alignment~\cite{lee2023aligning, wu2023better, kirstain2023pick, fan2024reinforcement, wallace2024diffusion}, or using AI feedback (RLAIF) from strong evaluation approaches or reward models~\cite{bai2022constitutional, yu2024rlaif}. 
Nevertheless, it is labor-intensive and costly to obtain extensive high-quality human feedback, which is also often required to train external reward models~\cite{black2023training}. 
Additionally, as LMMs evolve, the external evaluation approaches and reward models may require continual upgrading~\cite{pang2024language, yu2024rlaif}. 

To address the limitations, we consider utilizing LLMs' inherent discriminative capabilities to self-improve their generation quality for text-image alignment. 
This offers a pathway for LMMs to evolve for T2I independently, without relying on human or external feedback. 
To pursue self-improvement, the key steps are: 1) generating diverse images by LMMs based on a given prompt, ensuring the image diversity to facilitate subsequent self-assessment and optimization; 
2) using LMMs to self-assess text-image alignment in the generated images, producing alignment scores as self-feedback; 
and 3) adopting the self-feedback to optimize LMMs to generate superior visual tokens, resulting in images that better align with text prompts. 

However, achieving the above objectives faces significant challenges. In particular: 
\begin{itemize}
    \item [1)] As shown in Fig.~\ref{fig:intro}(b), LMMs typically generate intermediate visual representations, \ie, discrete visual tokens or continuous visual features, which are then converted into images by a decoder (\eg, a diffusion model)~\cite{sun2024generative, ge2024making}. 
    For LMMs with discrete visual tokens~\cite{yu2023scaling, ge2024making, wang2024emu3}, using existing sampling strategies (\eg, adjusting temperature) in the autoregressive generation process can obtain diverse visual tokens. 
    However, it is non-trivial for LMMs with deterministic continuous visual features, such as DreamLLM~\cite{dong2023dreamllm}, to sample diverse visual representations\footnote{Sampling diverse images at the decoder stage is inapplicable, as it can only optimize the decoder yet we aim to optimize LMMs to generate superior visual representations for text-image alignment in this work.}.

    \item [2)] Compositional prompts require LMMs to inspect object counts, attributes, and complex relationships in the generated images. However, existing LMMs still struggle with compositional cross-modal assessment~\cite{chen2024spatialvlm, rane2024can}, 
    challenging the generation of faithful self-feedback. 
    
    \item [3)] Optimizing LMMs with self-feedback is also intricate. Supervised Fine-Tuning (SFT)~\cite{dong2023abilities} and certain RLAIF methods~\cite{bai2022constitutional, yu2024rlaif} require highly accurate self-feedback. 
    Moreover, another representative method Direct Preference Optimization (DPO) requires modeling generation distributions, from which we need to sample diverse images to construct pairwise training data, which is challenging for LMMs with continuous visual features. 
    
\end{itemize}

\begin{figure*}[t]
\centering
	\includegraphics[width=1.01\textwidth]{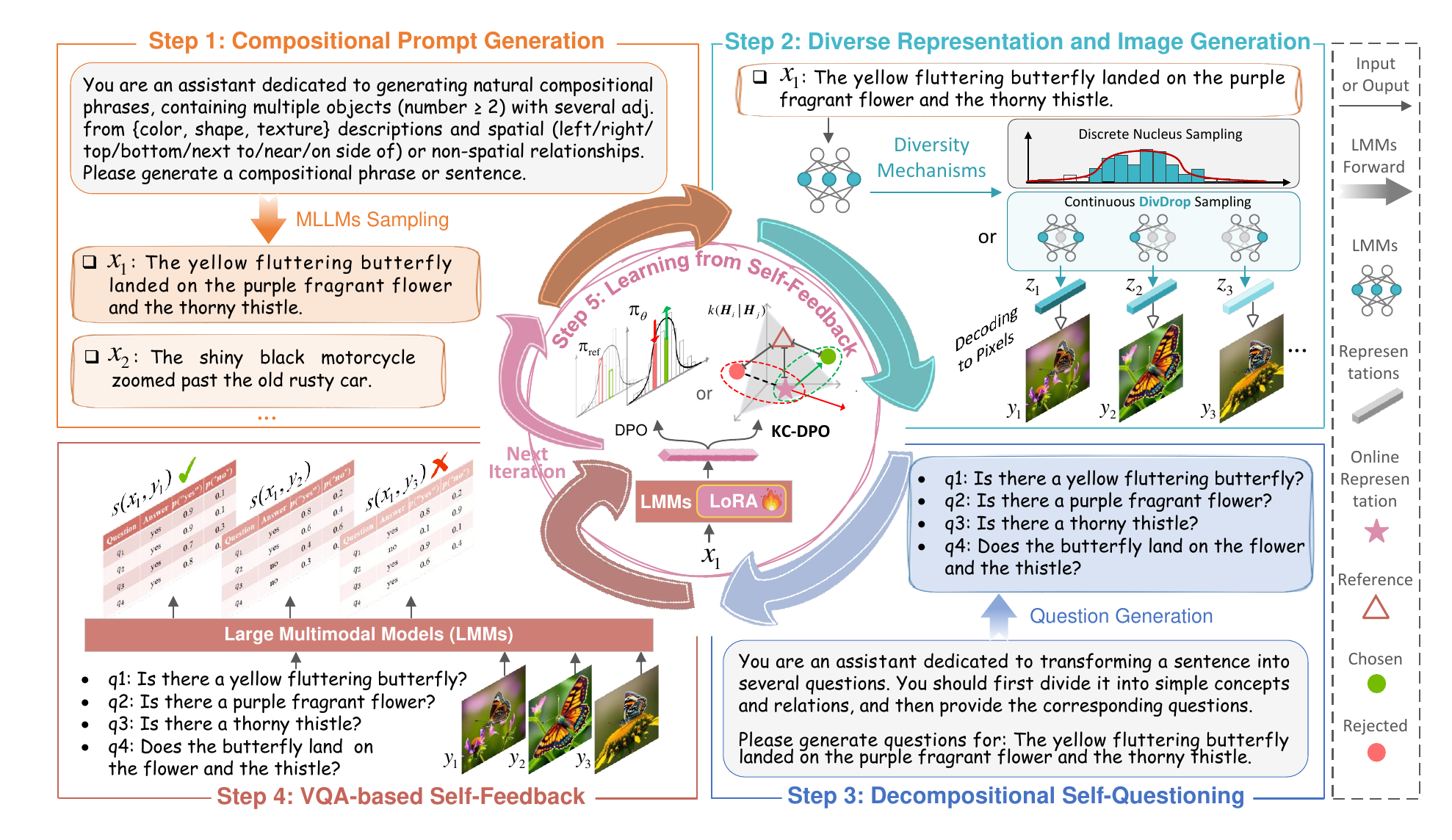}
	\vspace{-3ex}
	\caption{Schematic illustration of SILMM, comprising five steps: 1) LMMs generate compositional prompts by sampling based on provided instructions. 2) Diverse representations and images are generated using either discrete nucleus sampling or the proposed continuous DivDrop. 3) LMMs divide each compositional prompt into semantic units and generate questions for each unit. 4) VQA is conducted to answer these questions, with the answers and likelihoods aggregated into alignment scores as self-feedback. 5) For alignment tuning, DPO is applied for discrete LMMs, while the proposed KC-DPO is used for continuous LMMs.} 
	\label{fig:framework}
	\vspace{-2ex}
\end{figure*}

To tackle the above challenges, we propose an \underline{S}elf-\underline{I}mproving \underline{L}arge \underline{M}ultimodal \underline{M}odels (\textbf{SILMM}) framework for iterative optimization. 
As illustrated in Fig.~\ref{fig:framework}, SILMM operates through five steps: 1) \textbf{\textit{Compositional Prompt Generation}} prompts an LMM to imagine compositional scenarios and generate compositional prompts. 
2) \textbf{\textit{Diverse Image Generation}}. 
For discrete LMMs\footnote{For simplicity, we denote LLMs outputting discrete and continuous visual representations as discrete and continuous LMMs, respectively.}, we follow the sampling decoding strategy commonly used in LLM alignment~\cite{ouyang2022training, rafailov2024direct}. 
For continuous LMMs~\cite{dong2023dreamllm, wu2024next, sun2024generative}, we propose a diversification strategy named \textit{DropDiv}, inspired by Monte Carlo (MC) Dropout~\cite{gal2016dropout}, to perform dropout on the MLP layers of LMMs for diverse visual features, producing diverse images. 
3) \textbf{\textit{Decompositional Self-Questioning}}. 
To reduce the difficulty of compositional cross-modal assessment, LMMs can decompose a compositional prompt into atomic concepts and relations and generate questions for multi-step assessment. 
4) \textbf{\textit{VQA-based Self-Feedback}}. For each image generated in Step 2, LMMs can use the decomposed questions to assess text-image alignment, and then aggregate the results to obtain reasonable self-feedback. 
5) \textbf{\textit{Learning from Self-Feedback}}. 
For discrete LMMs, we directly apply DPO based on pairwise samples from Step 2. 
As to continuous LMMs, we propose Kernel-based Continuous DPO (\textbf{KC-DPO}), inducing a quadruplet objective with kernel functions for pairwise distance regulation over continuous visual features. 
The above five steps can iteratively repeat until self-improvement performance converges.

In summary, our main contributions are threefold: 
\begin{itemize}
    \item 
    To our knowledge, we are the first to focus on the task of LMMs' self-improvement for T2I. We propose a model-agnostic self-improvement framework to enable LMMs to achieve high-quality self-feedback and learning. 
    
    \item For continuous LMMs, we introduce a dropout-based strategy to diversify image representations, along with a continuous DPO approach, \ie, KC-DPO, to optimize LMMs with preference representation pairs. 
    
    \item We conduct extensive experiments on three compositional T2I benchmarks, demonstrating the superiority of SILMM, \eg, 30\% improvements on T2I-CompBench++. 
\end{itemize}

\section{Related Work}
\label{sec:related_work}
\mysubsubsec{Compositional Text-to-Image Generation}. 
Diffusion models~\cite{saharia2022photorealistic, ramesh2022hierarchical} have marked a significant advancement in T2I generation due to their stability and scalability. However, they still struggle with text-image alignment, such as attribute binding, counting error, and relation confusion~\cite{feng2023training, qu2024discriminative}. To enhance compositional T2I, some approaches intervene in language structures~\cite{feng2023training} or cross-attention mechanisms~\cite{chefer2023attend}. Other methods~\cite{qu2023layoutllm, feng2023layoutgpt, lian2023llm, lian2024llm} incorporate layout planning by LLMs or use multi-agent collaboration~\cite{wang2024divide, qin2024diffusiongpt}. Inspired by alignment successes in LLMs, recent work~\cite{black2023training, fan2024reinforcement, wallace2024diffusion} applies RLHF~\cite{ouyang2022training} to optimize diffusion models. Despite the progress, they rely on inductive biases, extensive prompt engineering, or labor-intensive annotations, limiting flexibility and scalability.

\mysubsubsec{Large Multimodal Models}. 
The pioneering LMMs~\cite{liu2023visual, zhu2023minigpt} integrate a visual encoder, \eg, CLIP~\cite{radford2021learning}, with LLMs as the foundation, showing impressive multimodal understanding capabilities. To extend LMMs to visual generation, recent approaches align diffusion models~\cite{dong2023dreamllm, wu2024next, ge2024making} with LLMs or train a single transformer~\cite{yu2023scaling, sun2024autoregressive, wang2024emu3, xie2024show}. According to the form of output visual features, they can be divided into discrete visual tokenization methods~\cite{ge2024making, sun2024autoregressive, wang2024emu3, qu2024unified} and continuous visual representation methods~\cite{dong2023dreamllm, wu2024next,sun2024generative}. 
While LLM integration enhances language understanding and supports flexible applications (e.g., interleaved multimodal generation~\cite{sun2024generative}), compositional T2I in the context of LMMs remains underexplored.

\mysubsubsec{Learning from AI Feedback}. 
The high cost of collecting human preference has spurred research into RLAIF~\cite{bai2022constitutional}. Benefiting from the convenience and scalability, there have been a series of studies adopting RLAIF to tackle a range of NLP tasks~\cite{lee2024rlaif, cui2024ultrafeedback, yuan2024self} and vision-language understanding~\cite{yu2024rlaif, wang2024rl}.
Despite the thrilling success, they only focus on text generation, overlooking the potential of RLAIF in other modalities. In contrast, we explore self-improving LMMs by activating multimodal understanding abilities for T2I. Particularly, we propose continuous strategies meticulously tailored to continuous visual features. 

\section{Methodology}\label{sec:method}
In this section, we elaborate on the proposed method, including the SILMM framework with five steps and the iteration strategy (Sec.~\ref{sec:silmm}), as illustrated in Fig.~\ref{fig:framework}. Afterward, we introduce the continuous KC-DPO applied to LMMs with continuous visual features in Sec.~\ref{sec:kc_dpo}. 

\subsection{Self-Improving Large Multimodal Models} \label{sec:silmm}
\mysubsubsec{Step 1: Compositional Prompt Generation}. 
We first divide compositional scenarios into four categories:  
\textit{Attribute} (color, shape, texture), \textit{Layout} (counting, spatial relation), \textit{Semantic Relation}, and \textit{Complex Composition}. Complex composition includes any possible composition of the first three. For attribute and layout, we prompt the LMM to separately generate common objects, attributes, numbers, and spatial relations, and then use templates to compose these concepts. For semantic relation and complex composition, we adopt in-context learning~\cite{dong2022survey} to generate prompts. More details can be found in App. ~\ref{sec:comp_prompt}.

\mysubsubsec{Step 2: Diverse Representation and Image generation}.
The purpose of this step is to sample diverse intermediate visual representations from the LLM backbone $\pi$ of an LMM, given a text prompt $x$, which would be decoded into images with different qualities. These representations are denoted as $\mathcal{Z} = \{z_i, ..., z_M\}$, where $z_i \sim \pi(z|x)$. 
For discrete LMMs~\cite{ge2024making}, $z_i$ is a discrete visual sequence. We follow the common practice~\cite{ouyang2022training, rafailov2024direct} in language generation to obtain $\mathcal{Z}$, by sampling with different random seeds during auto-regressive decoding. 
For continuous LMMs~\cite{dong2023dreamllm}, the LLM can only output a fixed continuous visual feature, without diversity. 
To tackle this issue, we propose \textit{DropDiv}. First, we insert the dropout operations in the last few MLP layers of LLMs, which introduces randomness and enables LLMs for sampling. During inference, we activate these dropout operations to output diverse representations by sampling: $z_i \sim \pi'(z|x)$, where $z_i$ denotes a continuous visual feature and $\pi'$ represents the LLM with activated dropout operations. 
Afterward, these diverse visual representations $\mathcal{Z}$ are decoded into images as $\mathcal{Y} = \{y_1, ..., y_M\}$. 

\mysubsubsec{Discussion}. 
Unlike prior work~\cite{black2023training, fan2024reinforcement} focused on tuning diffusion models, our approach resorts to LLM backbones in LLMs to control image decoders (\eg, diffusion models) for better text-image alignment, centering on LLM backbone optimization. Our approach offers three key advantages:
1) LLMs demonstrate superior proficiency in prompt comprehension over text encoders~\cite{raffel2020exploring, radford2021learning} commonly employed in diffusion models. Tuning LLM backbones may unlock their enormous potential for compositional T2I, especially in complex scenarios. 
2) Tuning diffusion models is often constrained by efficiency challenges inherent to iterative likelihood estimation, whereas there have been well-established  technologies~\cite{liu2020ipo, rafailov2024direct, azar2024general} for LLM alignment. 
3) Our method is orthogonal to existing methods to tune diffusion models, combining them may get further gains.

\mysubsubsec{Step 3: Decompositional Self-Questioning}. 
To provide helpful feedback to the generated images, the LMM should first accurately assess text-image alignment, which requires strong compositional reasoning abilities. However, current advanced LMMs still suffer from compositional reasoning~\cite{mitra2024compositional}, such as spatial relation understanding~\cite{chen2024spatialvlm} and counting~\cite{rane2024can}. 
To improve compositional reasoning, we introduce a divide-and-conquer strategy~\cite{yu2024rlaif} for self-questioning. 
Specifically, the LMM first divides the given prompt $x$ into atomic concepts (\eg, ``a white harp'') and relations (\eg, ``a pancake is on the left of a pasta''), and then generates questions $\mathcal{Q} = \{q_1, ..., q_N\}$, each $q_i$ corresponding to a concept or relation. 
For simplicity, the generated questions are constrained to be yes/no questions (\eg, ``Is there a while harp?'', ``Is the pancake on the left of the pasta?''). Refer to App.~\ref{sec:self_q_prompt} for more details on prompt templates of self-questioning. 

\mysubsubsec{Step 4: VQA-based Self-Feedback}. 
Taking a generated image $y \in \mathcal{Y}$ and all the questions $\mathcal{Q}$ as input, the LMM conducts the VQA task, and the average difference between the probabilities of answering ``yes'' and ``no'' serves as the text-image alignment score: 
\begin{equation}\label{eqn:alignment_score}
    s(x, y) = \frac{1}{N} \sum_{i=1}^{N} [p(``yes"|y, q_i) - p(``no"|y, q_i)]. 
\end{equation}
Here we adopt the vision-language understanding abilities of LMMs via VQA to provide feedback to the images generated by themselves, thus this step is named VQA-based self-feedback. We carry out this step through all the sampled images prompted by $x$ and get all the scores $\mathcal{S} = \{s(x, y_j) | y_j \in \mathcal{Y}\}$. 

\mysubsubsec{Step 5: Learning from Self-Feedback}. 
Based on the self-feedback alignment scores, we sample representation pairs $(z_w, z_l)$ from $\mathcal{Q}$, where $z_w$ and $z_l$ denote the chosen and the rejected representations and their corresponding decoded images should satisfy $s(x, y_w) > s(x, y_l)$. 
With the preference data, we optimize the LLM backbone with DPO~\cite{rafailov2024direct}: 
\begin{multline}\label{eqn:dpo}
\small 
    \mathcal{L}_{\text{DPO}} = -\mathbb{E}_{(x, z_w, z_l) \sim \mathcal{D}} \\
    \left[\log \sigma \left(\beta \log \frac{\pi_\theta (z_w | x)}{\piref (z_w | x)} - \beta \log \frac{\pi_\theta (z_l | x)}{\piref (z_l | x)}\right)\right], 
\end{multline}
where $\mathcal{D}$ denotes the training set, and $\pi_\theta$ and $\piref$ represent the policy and reference models, respectively. $\sigma$ is the sigmoid function, and $\beta$ is a hyperparameter controlling the deviation from the reference model. 

\mysubsubsec{Iterative Self-Improvement}. 
After learning from self-feedback, the updated LMM becomes more likely to generate preferred representations that are decoded into images better aligned with the prompt. This improvement in overall text-image alignment motivates us to iterate the above five steps with the updated LMM as the new reference model. The iteration mechanism continues until the alignment performance converges. As the process is independent of human annotations and external models, it is cost-effective and scalable. More importantly, it showcases the potential for self-improvement in LMMs by harmonizing their understanding and generation capabilities.

\subsection{Continuous Direct Preference Optimization}
\label{sec:kc_dpo}
At the step of learning from self-feedback, LMMs are optimized using the DPO objective as shown in Eqn.~(\ref{eqn:dpo}). 
The difference between discrete and continuous LMMs in this learning process lies in the calculation of the likelihood $\pi(z | x)$. For discrete LMMs, $\pi(z | x)$ can be straightforwardly obtained by the softmax categorical distribution. However, for continuous LMMs with unknown distribution modeling, calculating $\pi(z | x)$ is intractable. 

\mysubsubsec{Predictive Distribution with MC Dropout}. 
MC Dropout~\cite{gal2016dropout} enables predictive distribution estimation via Monte Carlo simulation to calculate $\pi(z | x)$. Specifically, the dropout layers\footnote{In fact, there is no dropout layer in most open-sourced LLMs (\eg, LLaMA series~\cite{touvron2023llama, touvron2023llama2, dubey2024llama}), and a compromise solution is to introduce additional dropout layers.} in an LMM are activated during inference and the LMM performs forward propagation multiple times to get multiple outputs. Assuming a Gaussian distribution, we can estimate its parameters and calculate the likelihood $\pi(z | x)$ based on these outputs. However, such multi-forward estimation imposes a significant computational burden during training, making this approach insufficient and impractical. 

\mysubsubsec{Simplified Kernel-based Continuous DPO}. 
Inspired by MC Dropout and motivated by its insufficiency issue, we propose a simplified method to achieve continuous DPO. 
Concretely, the intermediate representation $z$ often performs as a feature matrix $\bm{H} \in \mathbb{R}^{L \times D}$ where $L$ and $D$ denote the sequence length and dimension. $\bm{H}$ can be attained by a Q-Former~\cite{dong2023dreamllm, ge2024seed} or from the last layer of the LMM in an autoregressive way~\cite{sun2023emu, sun2024generative}.
To estimate $\pi(\bm{H} | x)$, we first make a decomposition as:
\begin{equation}\label{eqn:decomp}
    \pi(\bm{H} | x) = \prod_{i=1}^{L} \pi(\bm{h}_i|\bm{H}_{<i}, x), 
\end{equation}
where $\bm{h}_i \in \mathbb{R}^D$ denotes the $i$-th feature vector. Based on the Gaussian assumption, we have: 
\begin{equation}\label{eqn:gau}
    \small 
    \pi(\bm{h}_i|\bm{H}_{<i}, x) = \frac{\exp\left[-\frac{1}{2}(\bm{h}_i-\bm{\mu}_i)^\top\bm{\Sigma}_i^{-1}(\bm{h}_i-\bm{\mu}_i)\right]}{\sqrt{(2\pi)^D |\bm{\Sigma}_i|}}, 
\end{equation}
where $\bm{\mu}_i$ and $\bm{\Sigma}_i$ denote the mean vector and the covariance matrix, respectively. Furthermore, we further simplify and approximate this formula: 1) the mean vector is estimated by the direct output of the continuous LMM, \ie, $\bm{\mu}_i \approx {\rm LMM}(x)[i] $, and 2) the Gaussian distribution is isotropic and all dimensions share the same variance value $\bar{\sigma}$, \ie, $\bm{\Sigma}_i \approx \rm{diag} (\sigma_1, ..., \sigma_D)$ and $\sigma_1 = ... = \sigma_D = \bar{\sigma}$, and $\bar{\sigma}$ can be learnable or viewed as a hyperparameter. 

We compute the simplified likelihood with Eqn.~(\ref{eqn:gau}), obtain the joint one with Eqn.~(\ref{eqn:decomp}), and finally derive the continuous DPO based on Eqn.~(\ref{eqn:dpo}):
\begingroup
\small 
\setlength{\jot}{-5pt} 
\begin{align}\label{eqn:cont_dpo}
    & \mathcal{L}_{\text{C-DPO}} = -\mathbb{E}_{(x, \bm{H}_w, \bm{H}_l) \sim \mathcal{D}} 
    \Bigg[\log\sigma \bigg(\frac{\beta}{2\bar{\sigma}^2} 
    (-\Vert \bm{H} - \bm{H}_w \Vert_F^2 \nonumber \\
    & \quad + \Vert \bm{H}_r - \bm{H}_w \Vert_F^2 
     + \Vert \bm{H} - \bm{H}_l \Vert_F^2 
     - \Vert \bm{H}_r - \bm{H}_l \Vert_F^2 )\bigg) \Bigg], 
\end{align}
\endgroup
where $\Vert \cdot \Vert_F$ denotes the Frobenius norm, and $\bm{H}$ and $\bm{H}_r$ represent the continuous feature matrices from the policy and reference LMMs, respectively. $\bm{H}_w$ and $\bm{H}_l$ refer to the chosen and rejected feature matrices, respectively. Compared with the MC dropout method, this objective only requires one forward pass, which is more efficient. 
We relegate more details of the derivation to App.~\ref{sec:kc_dpo_deriv}. 
From Eqn.~(\ref{eqn:cont_dpo}), we can see that this objective aims to adjust the relative distances within the quadruple $(\bm{H}, \bm{H}_r, \bm{H}_w, \bm{H}_l)$ and the distance metric is the Euclidean distance between two matrices. To further improve the flexibility, we generalize the continuous DPO objective to, 
\begingroup
\small 
\setlength{\jot}{-5pt} 
\begin{align}\label{eqn:kc_dpo}
    \mathcal{L}_{\text{KC-DPO}} &= -\mathbb{E}_{(x, \bm{H}_w, \bm{H}_l) \sim \mathcal{D}} 
    \Bigg[\log\sigma \bigg( \gamma 
    (-k(\bm{H}, \bm{H}_w) \nonumber \\
    & \quad + k(\bm{H}_r, \bm{H}_w) 
     + k(\bm{H}, \bm{H}_l)  
     - k(\bm{H}_r, \bm{H}_l) ) \bigg) \Bigg], 
\end{align}
\endgroup
where $\gamma = \frac{\beta}{2\bar{\sigma}^2}$ controls the degree of adherence to the reference model, $k(\cdot, \cdot)$ denotes a generalized distance measurement function. Considering it is similar to kernel methods~\cite{shawe2004kernel, hearst1998support}, we name the objective Kernel-based Continuous DPO (KC-DPO). In the following experiments section, we will discuss different distance functions and their influences on alignment performance. 

\section{Experiments}\label{sec:exp}
\subsection{Experimental Setup}
\mysubsubsec{Base Model Settings.} We implement our method on DreamLLM (continuous LMM)~\cite{dong2023dreamllm} and SEED-LLaMA (discrete LMM)~\cite{ge2024making} for all experiments. We also apply our method to Emu-3~\cite{wang2024emu3}, the recent state-of-the-art discrete LMM. Details on DPO training are provided in App.~\ref{sec:app_dpo_train}. 

\mysubsubsec{Datasets.}
We curated a dataset of 16,000 prompts across four categories using LMM. In each DPO training iteration, images generated by the model in the previous iteration served as the training data for next DPO iteration, allowing for iterative self-improvement. Details on data creation are provided in App.~\ref{sec:dpo_data}. 

\mysubsubsec{Benchmarks.}
We evaluate our method on three text-to-image alignment benchmarks and follow their default settings. T2I-CompBench++~\cite{huang2023t2i} consists of 8,000 compositional text prompts organized into 4 main categories: attribute, layout, non-spatial, and complex compositions, further divided into 8 subcategories, including color, binding, binding, 2D/3D-spatial relationships, non-spatial relationships, numeracy, and complex compositions. TIFA~\cite{hu2023tifa} uses pre-generated question-answer pairs and a VQA model to evaluate generation results based on 4,081 diverse text prompts and 25,829 questions across 12 categories. DPG-Bench~\cite{hu2024ella} comprises 1,065 densely descriptive prompts with an average token length of 83.91, presenting more complex scenarios with varied objects and rich adjectives.

\subsection{Performance Comparison}
\begin{table*}[t]
	\centering
	\setlength{\abovecaptionskip}{0.15cm}
	\caption{Performance comparison and improvement of the proposed method for compositional text-to-image generation on T2I-CompBench++~\cite{huang2023t2i}, DPG-Bench~\cite{hu2024ella}, and TIFA~\cite{hu2023tifa}. Alignment scores are calculated using expert understanding models (\eg, VQA or object detection models) recommended by these benchmarks. Prompt rewriting in Emu3~\cite{wang2024emu3} was not used for fair comparison. 
 }
	\label{tab:perf_comp}
	{\hspace{-1ex}
             \setlength{\tabcolsep}{1mm}{
		\resizebox{0.87\textwidth}{!}
		{
			\setlength\tabcolsep{5pt}
			\renewcommand\arraystretch{1.1}
                \begin{tabular}{l|cccc|cccccc|c}
				\hline\thickhline
				\multicolumn{1}{c|}{\multirow{2}{*}{Method}} & \multicolumn{4}{c|}{ T2I-CompBench++~\cite{huang2023t2i}}  & \multicolumn{6}{c|}{DPG-Bench~\cite{hu2024ella}} & \multicolumn{1}{c}{TIFA~\cite{hu2023tifa}} \\ 
				\cline{2-12}
				 & Attribute & Layout  & Non-spatial & Complex & Global & Entity & Attribute  & Relation & Other & All & All \\
				\hline\hline
                \multicolumn{12}{l}{\textit{Text-to-Image Generative Models}} \\
                \hline
                \multicolumn{1}{l|}{SD-v1.5~\cite{ramesh2022hierarchical}} & 38.65  & -     & -     & -     & 74.63  & 74.23  & 75.39  & 73.49  & 67.81  & 63.18  & 78.40  \\
                \multicolumn{1}{l|}{DALL-E 2~\cite{ramesh2022hierarchical}} & 58.63  & -     & -     & -     & -     & -     & -     & -     & -     & -     & - \\
                \multicolumn{1}{l|}{SD-v2~\cite{ramesh2022hierarchical}} & 47.36  & 30.50  & 31.27  & 33.86  & 77.67  & 78.13  & 74.91  & 80.72  & 80.66  & 68.09  & - \\
                \multicolumn{1}{l|}{SD-v2.1~\cite{ramesh2022hierarchical}} & 50.57  & -     & -     & -     & -     & -     & -     & -     & -     & -     & 82.00  \\
                \multicolumn{1}{l|}{SDXL~\cite{podell2023sdxl}}  & 52.88  & 35.62  & 31.19  & 32.37  & 83.27  & 82.43  & 80.91  & 86.76  & 80.41  & 74.65  & - \\
                \multicolumn{1}{l|}{PixArt-$\alpha$~\cite{chen2023pixart}} & 60.31  & 36.74  & 31.97  & 34.33  & 74.97  & 79.32  & 78.60  & 82.57  & 76.96  & 71.11  & - \\
                \multicolumn{1}{l|}{DALL-E 3~\cite{betker2023improving}} & 70.09  & 41.63  & 30.03  & 37.73  & 90.97  & 89.61  & 88.39  & 90.58  & 89.83  & 83.50  & - \\
                \hline
                \multicolumn{12}{l}{\textit{Large Multimodal Models}} \\
                \hline
				\multicolumn{1}{l|}{SEED-LLaMA~\cite{ge2024making}} & 19.20  & 20.29  & 28.86  & 21.46  & 65.59 & 55.87 & 61.96 & 62.77 & 59.46 & 47.12 & 66.74  \\
                \multicolumn{1}{l|}{SEED-LLaMA + Ours} & 39.60  & 25.11  & 29.82  & 28.28  & 73.55 & 70.48 & 68.49 & 74.79 & 68.64 & 57.31  & 73.74  \\
                \rowcolor{gray_bg}
                \multicolumn{1}{l|}{\%Improvment} & 106.25\% & 23.77\% & 3.33\% & 31.78\% & 12.14\% & 26.15\% & 10.54\% & 19.15\% & 15.44\% & 21.63\% & 10.49\% \\
                \multicolumn{1}{l|}{DreamLLM~\cite{dong2023dreamllm}} & 22.94  & 23.74  & 28.76  & 23.01  & 74.47  & 65.86  & 63.80  & 74.24  & 46.00  & 53.93  & 69.91  \\
                \multicolumn{1}{l|}{DreamLLM + Ours} & 39.94  & 27.63  & 29.00  & 26.43  & 76.29  & 75.91  & 69.20  & 84.41  & 60.00  & 64.22  & 75.38  \\
                \rowcolor{gray_bg}
                \multicolumn{1}{l|}{\%Improvment} & 74.15\% & 16.40\% & 0.83\% & 14.86\% & 2.44\% & 15.26\% & 8.46\% & 13.70\% & 30.43\% & 19.08\% & 7.82\% \\
                \multicolumn{1}{l|}{Emu3~\cite{wang2024emu3}} & 44.79  & 32.30  & 30.15  & 31.32  & 84.19  & 80.81  & 82.75 & 87.23 & 50.80 & 74.19 & 81.86 \\
                \multicolumn{1}{l|}{Emu3 + Ours} & 59.71  & 36.03  & 30.51  & 33.93  & 84.19  & 81.57  & 84.52  & 89.01  & 64.80  & 77.45  & 85.11  \\
                \rowcolor{gray_bg}
                \multicolumn{1}{l|}{\%Improvment} & 33.30\% & 11.57\% & 1.19\% & 8.33\% & 0.00\% & 0.94\% & 2.14\% & 2.04\% & 27.56\% & 4.39\% & 3.97\% \\
				\hline
			\end{tabular}
		}
            }
	}
\vspace{-2ex}
\end{table*}

As shown in Tab.~\ref{tab:perf_comp}, we evaluate alignment performance of our method against T2I generative models and base LMMs on three compositional T2I benchmarks, including T2I-CompBench++~\cite{huang2023t2i}, DPG-Bench~\cite{hu2024ella}, and TIFA~\cite{hu2023tifa}. 
Key observations are as follows: 1) Although LMMs enable more flexible settings (\eg, in-context learning and interleaved multimodal generation) for image generation, they still underperform compared to specialized T2I models in terms of the basic alignment ability to follow prompts. It demonstrates that current LMMs may ignore the compositional text-image alignment during multimodal pre-training and fine-tuning. 2) Without human annotations or external models, the proposed SILMM method enhances alignment performance across all categories in three benchmarks over the base LMMs, improving both the discrete SEED-LLaMA and the continuous DreamLLM, verifying the effectiveness and the generalization of SILMM. 3) SEED-LLaMA shows greater self-improvement than DreamLLM, possibly due to its weaker baseline alignment and the stability of discrete DPO over continuous KC-DPO induced by a series of simplification, as discussed in Sec.~\ref{sec:kc_dpo}. 
And 4) improvements are more challenging in layout, relation, and complex categories than in attribute categories. This difficulty arises partly because the basic generative ability in these categories is weak, making it difficult to obtain high-quality chosen samples. Besides, understanding compositional concepts remains a challenge for LMMs~\cite{chen2024spatialvlm, rane2024can}.

\subsection{In-depth Analysis}
To explore the efficacy of SILMM, we conduct extensive ablation studies and hyperparameter analyses. We first investigate the iteration process and data scaling, followed by an in-depth study of key components, including diversity strategies, decompositional self-questioning and answering for self-feedback, and KC-DPO. 

\begin{figure}[t]
        \centering
	\includegraphics[width=0.35\textwidth]{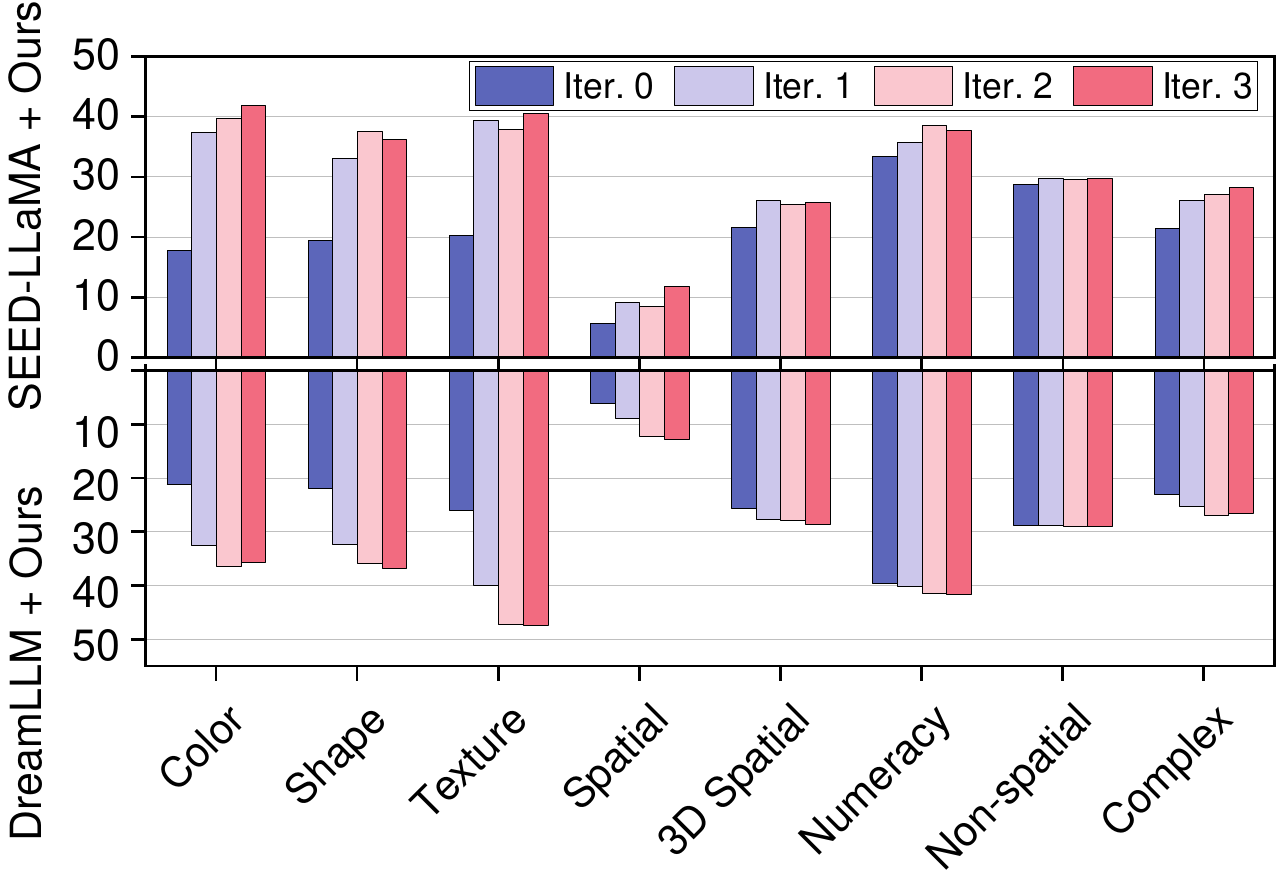}
	\vspace{-2ex}
	\caption{Performance improvement of iterative alignment tuning based on SEED-LLaMA and DreamLLM, across 8 detailed categories of T2I-CompBench++. Iter. 0 denotes the base models without alignment tuning. }
	\label{fig:iter}
	\vspace{-3ex}
\end{figure}

\mysubsubsec{Iterative Self-Improvment}. 
As shown in Fig.~\ref{fig:iter}, we conduct three iterations of self-improvement and assess performance changes across eight detailed categories of T2I-CompBench++~\cite{huang2023t2i}. The results show that SILMM achieves effective, consistent, and continuous improvements in text-image alignment, across most compositional categories. Notably, attribute categories (\eg, color, shape, and texture) exhibit the most significant gains, whereas the non-spatial category shows slower improvement. This slower progress may stem from CLIP score~\cite{hessel2021clipscore}, which is less sensitive than other metrics. Finally, as the iteration progresses, improvement rates gradually decrease, indicating convergence. More iterative self-improvement experiments results can be found in App.~\ref{sec:add_exp}.

\begin{figure}[h]
\centering
 \subfloat[Num. of Training Prompts]{
\label{fig:prompt_num}
\includegraphics[scale=0.19]{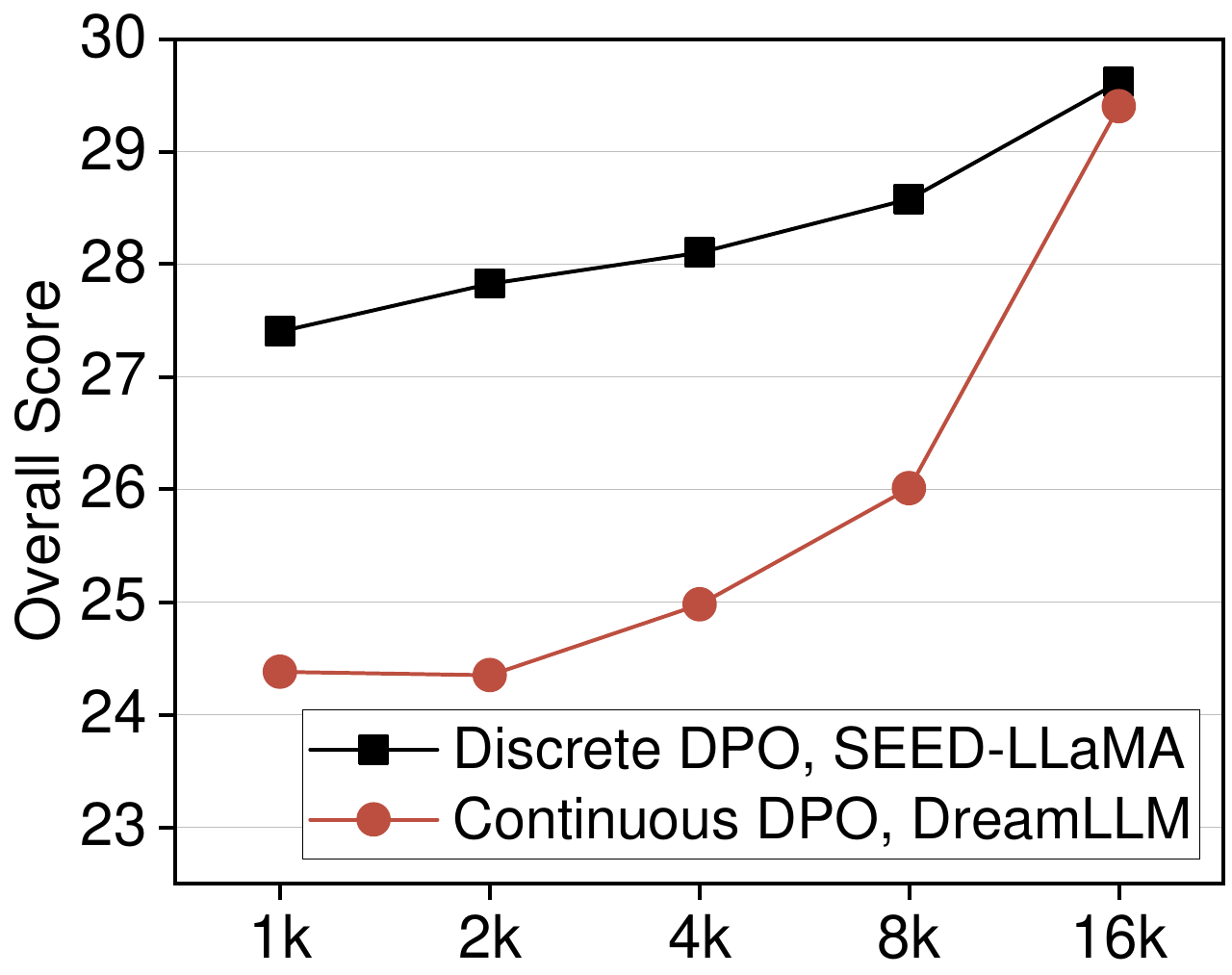}}\hspace{-0.15cm}
\subfloat[Num. of Pairs per Prompt]{
\label{fig:pair_num}
\includegraphics[scale=0.179]{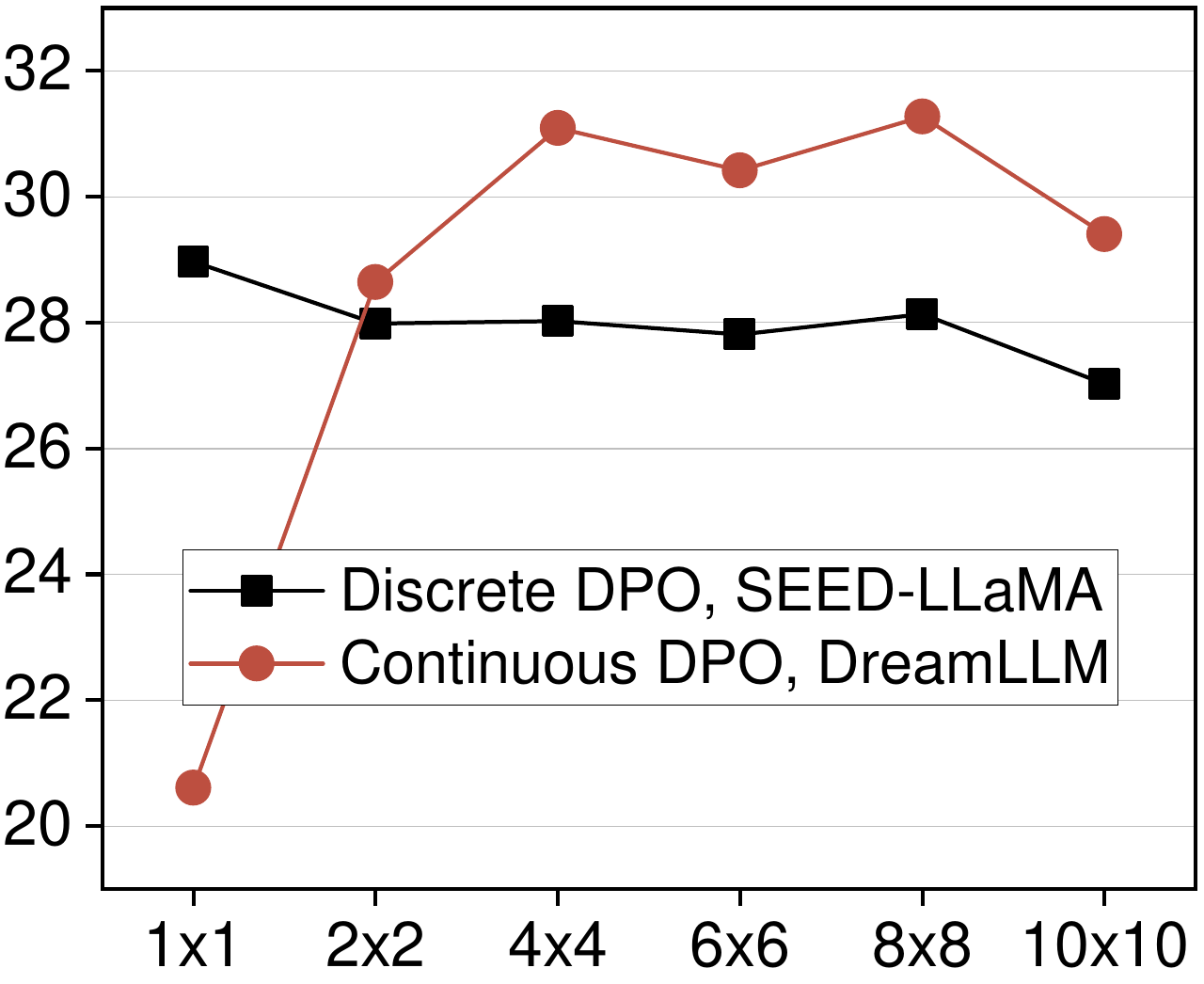}}\vspace{-0.1cm} 
\vspace{-1ex}
\caption{Overall alignment scores of SEED-LLaMA with discrete DPO and DreamLLM with continuous KC-DPO, on T2I-CompBench++ with (a) varying numbers of generated prompts in the training data, and (b) different number of preference pairs sampled from 30 diverse generated images per prompt. $N \times N$ means we select the top-N and last-N images from 30 generated ones as the chosen and rejected, respectively. }
\label{fig:data_scale}
\vspace{-2ex}
\end{figure}
\mysubsubsec{Data Scale}. 
The proposed SILMM method leverages self-synthesized data for tuning, allowing flexible adjustment of data scale according to practical needs and available computational resources. In Fig.~\ref{fig:data_scale}, we investigate how data scale affects overall alignment performance (averaged across eight categories in T2I-CompBench++), focusing on two factors: the number of training prompts and the number of preference pairs per prompt. Results in Fig.~\ref{fig:prompt_num} indicate that both LMMs show consistent improvement as data samples increase, showing the strong scalability of the proposed method. Besides, we generate 30 representations and images per prompt, and then select the top-$N$ and last-$N$ samples to construct $N \times N$ preference pairs (see Fig.~\ref{fig:pair_num}). Notably, the two LMMs perform differently. This may be because the continuous feature space, being larger and denser than the discrete space, requires denser data pairs to stabilize the optimization dynamics.

\begin{figure*}[h]
\centering
 \subfloat[Color, Shape, and Texture]{
\label{fig:div_attr}
\includegraphics[scale=0.21]{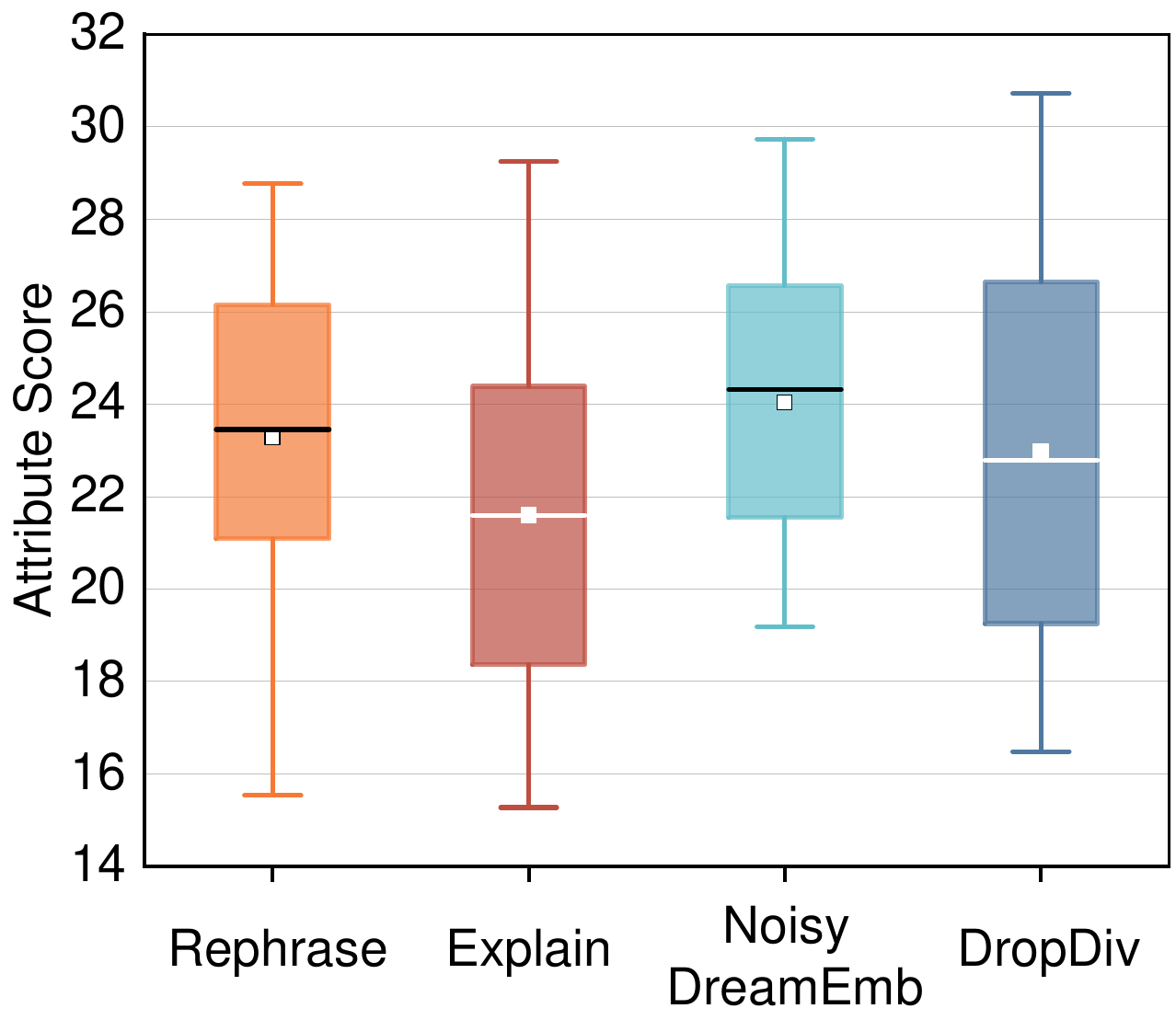}}\hspace{0.4cm}
\subfloat[Spatial, 3D Spatial, and Numeracy]{
\label{fig:div_layout}
\includegraphics[scale=0.21]{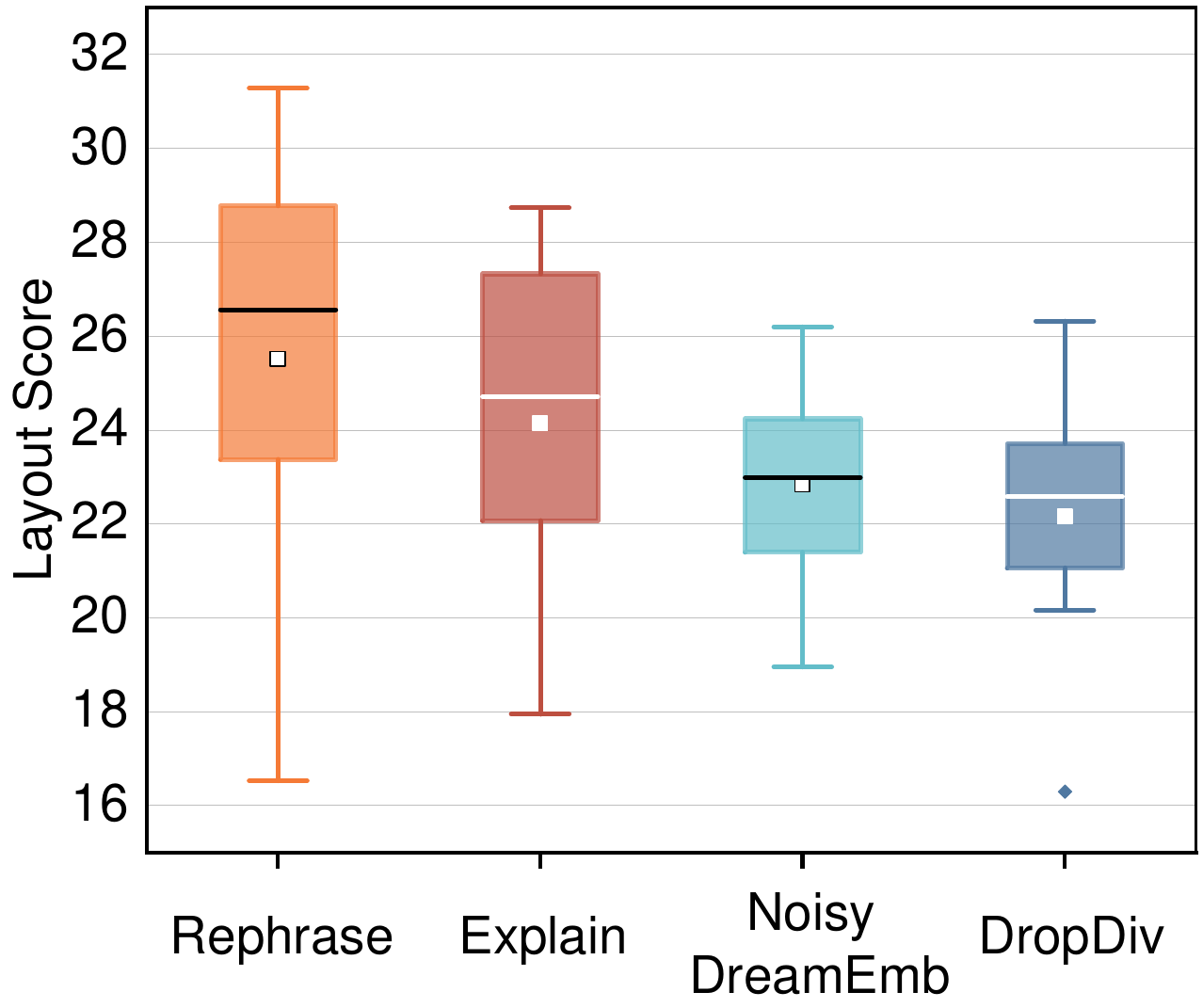}}\hspace{0.4cm}
\subfloat[Non-spatial and Complex]{
\label{fig:div_nonspa_complex}
\includegraphics[scale=0.215]{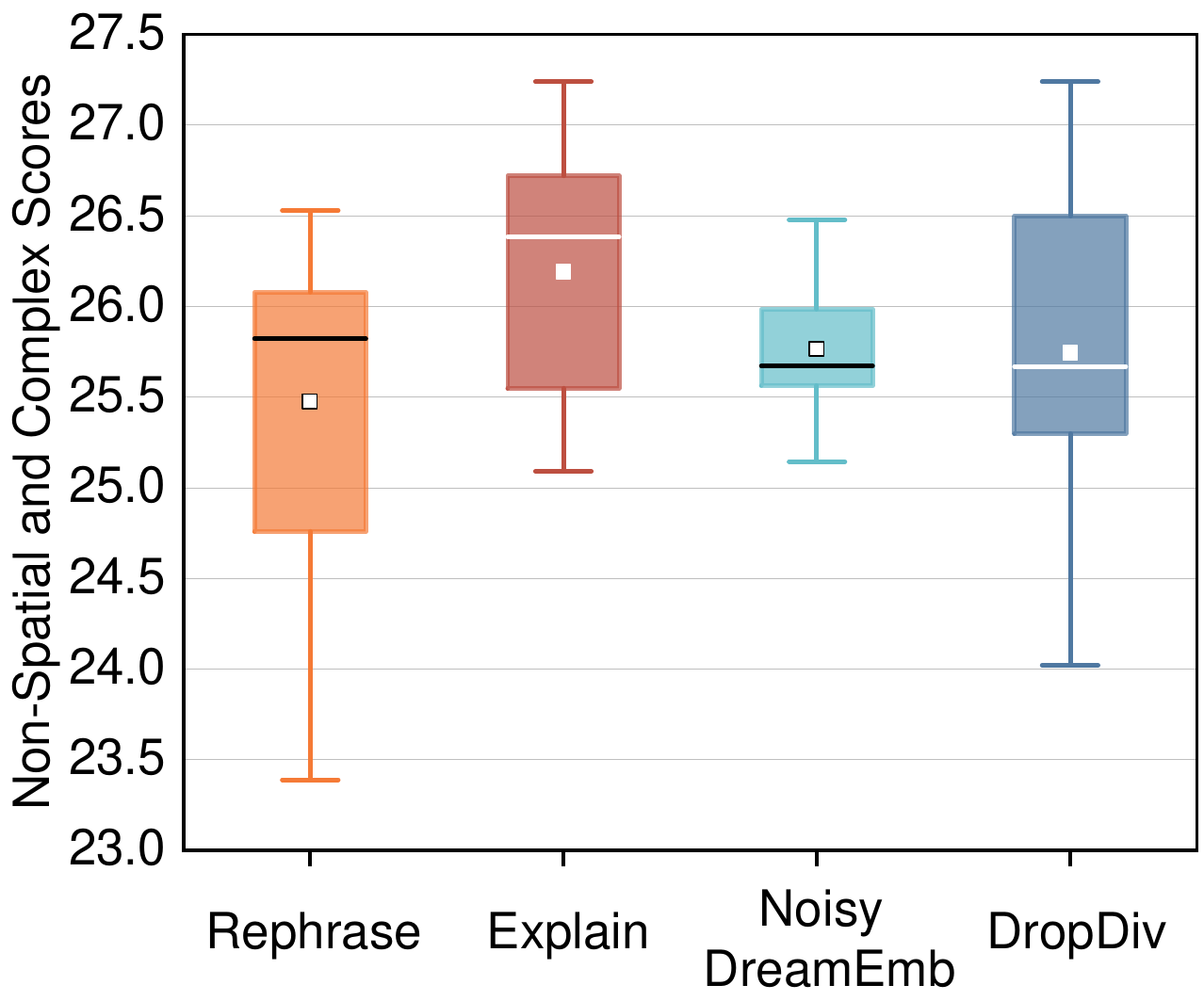}}
\vspace{-1ex}
\caption{Comparison of four methods for diverse continuous representation generation, with alignment scores evaluated on the validation set of T2I-CompBench++. For each prompt, DreamLLM generates ten diverse representations and corresponding images. }
\label{fig:div}
\vspace{-2ex}
\end{figure*}

\begin{figure}[t]
        \centering
	\includegraphics[width=0.40\textwidth]{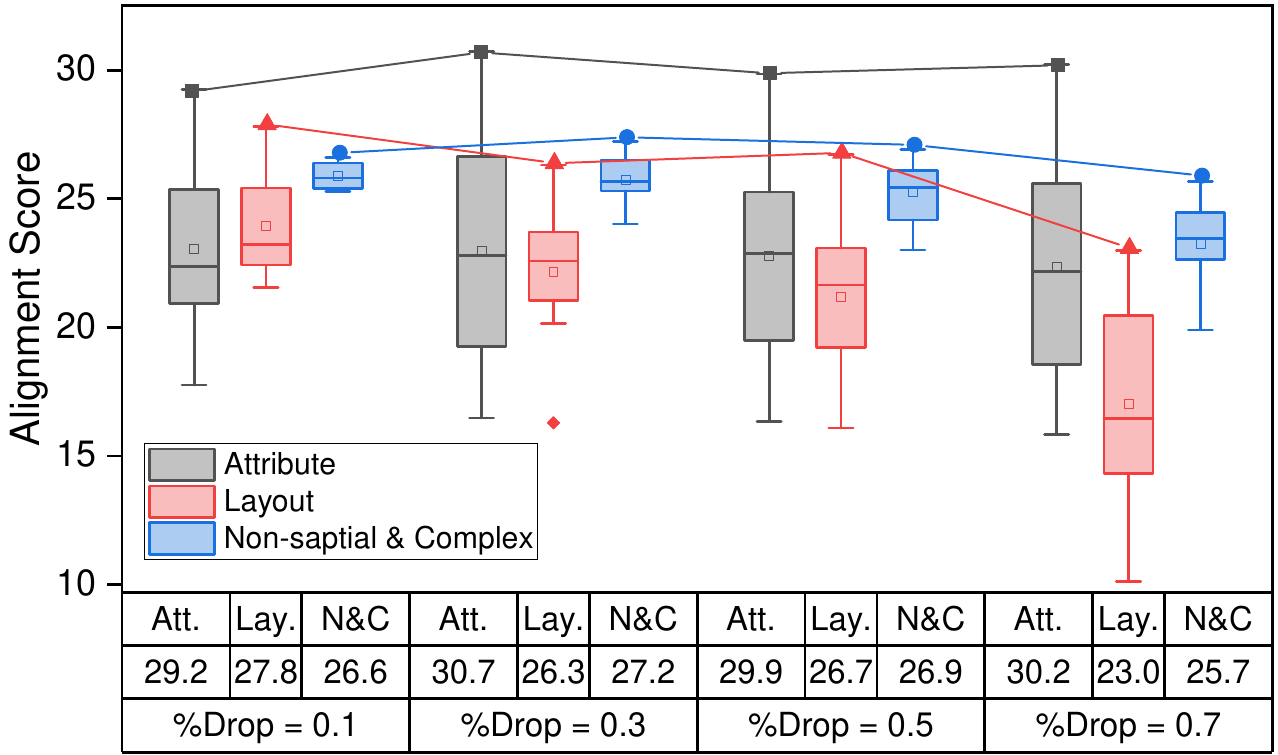}
	\caption{Distribution of alignment scores and variation of maximum scores across different dropout rates (\%Drop) of the proposed DropDiv method, evaluated on the T2I-CompBench++ validation set. For each prompt, DreamLLM generates ten diverse representations and corresponding images with DropDiv. }
	\label{fig:drop_rate}
	\vspace{-3ex}
\end{figure}
\mysubsubsec{Diversity Strategies}. 
To synthesize high-quality preference pair data, the diversity strategy is crucial. An effective diversity strategy should maximize the potential of LMMs while ensuring sufficient variation among generated images. 
To explore various strategies, we compare the proposed \textit{DropDiv} with three alternatives: \textit{Rephrase} the original prompt, \textit{Explain} the original prompt with added imaginative elements, and add Gaussian noises to the learnable Dream Embeddings in DreamLLM~\cite{dong2023dreamllm} to create \textit{Noisy DreamEmb}. 
As shown in Fig.~\ref{fig:div}, four strategies have different perturbation influences on different categories. For example, DropDiv could generate better samples in attribute, non-spatial, and complex categories, but compromise in layout categories. To further examine the effects of DropDiv, we conduct experiments across different dropout rates as shown in Fig.~\ref{fig:drop_rate}. Results indicate that higher dropout rates enhance diversity, but the alignment quality could be impaired. Therefore, achieving a good diversity-quality trade-off remains challenging.

\begin{table}[t]
	\centering
	\setlength{\abovecaptionskip}{0.15cm}
	\caption{Ablation study on T2I-CompBench++~\cite{huang2023t2i} and DPG-Bench~\cite{hu2024ella} examining variations in \textbf{Question Generation} and \textbf{VQA-based Alignment Score Calculation} methods for self-feedback. Prompt-Q adds a ``?'' or replaces the period with a ``?'' at the end of each prompt. Phrase-Q involves dividing a prompt into phrases, each followed by a ``?''. Self-Q instructs the LMM to generate questions for each prompt using in-context examples. Diff. of Prob. denotes the proposed alignment score calculation approach described in Eqn.~(\ref{eqn:alignment_score}). 
 }
	\label{tab:abla_question_gen}
	{\hspace{-1ex}
             \setlength{\tabcolsep}{1mm}{
		\resizebox{0.40\textwidth}{!}
		{
			\setlength\tabcolsep{3pt}
			\renewcommand\arraystretch{1.1}
                \begin{tabular}{c|cccc|c}
				\hline\thickhline
                \multicolumn{1}{c|}{\multirow{2}{*}{Feedback}} & \multicolumn{4}{c|}{ T2I-CompBench++}  & \multicolumn{1}{c}{DPG-Bench}  \\ %
                \cline{2-6}
                  & Attribute & Layout  & Non-spatial & Complex & All\\
				\hline\hline
                \multicolumn{6}{l}{\textit{Baseline (DreamLLM~\cite{dong2023dreamllm})}} \\
                -  & 22.94 & 23.74  & 28.76 & 23.01 & 53.93 \\
                \hline
                \multicolumn{6}{l}{\textit{Question Generation}} \\
                \hline
                Prompt-Q & 33.66 & \textbf{25.93}  & 28.14 & 24.67  & 60.13  \\
                Phrase-Q & 34.63 & 24.91  & 28.01 & \textbf{25.41} & 60.10   \\
                \rowcolor{gray_bg}
                Self-Q & \textbf{34.85} & 25.51   & \textbf{28.82} & 25.31  & \textbf{60.95}  \\
                \hline
                \multicolumn{6}{l}{\textit{VQA-based Alignment Score Calculation}} \\
                \hline
                Random Score & 23.41 & 24.81  & 28.67 & 22.95  & 53.89 \\
                Ratio of ``yes'' & 25.36 & 23.51  & 28.73 & 24.00 & 54.68 \\
                \rowcolor{gray_bg}
                Diff. of Prob. & \textbf{34.85} & \textbf{25.51} & \textbf{28.82} & \textbf{25.31}  &\textbf{ 60.95} \\
                \hline
			\end{tabular}
		}
            }
	}
\hspace{-3ex}
\end{table}

\mysubsubsec{Decompositional Self-Feedback}. 
We perform ablation studies on question generation and alignment score calculation, as shown in Tab.~\ref{tab:abla_question_gen}. Compared to two variants \textit{Prompt-Q} (appending or replacing the period with ``?'' at the end of each prompt) and \textit{Phrase-Q} (segmenting each prompt into phrases using NLP tools~\cite{honnibal2020spacy}), Self-Questioning (Self-Q) achieves better alignment performance across most categories, demonstrating the effectiveness of leveraging language processing abilities of LMMs for text-image alignment evaluation. Additionally, we compare the proposed alignment score calculation from Eqn.~(\ref{eqn:alignment_score}) with two variants: \textit{Random Score} and \textit{Ratio of ``yes''} (where a higher ratio indicates a higher score). Results show that our method achieves superior performance by considering the relative confidence between ``yes'' and ``no''. 

\begin{table}[t]
	\centering
	\setlength{\abovecaptionskip}{0.15cm}
	\caption{Ablation study on T2I-CompBench++~\cite{huang2023t2i} to investigate different instantiation of the \textbf{Kernel Function} to calculate the continuous KC-DPO loss function to tune DreamLLM. Aggregation means we aggregate the 2D feature matrix (\eg, $\bm{H}$) into 1D along the sequence dimension. Eucl. denotes Euclidean distance. 
 }
	\label{tab:abla_kernel}
	{\hspace{-1ex}
             \setlength{\tabcolsep}{1mm}{
		\resizebox{0.38\textwidth}{!}
		{
			\setlength\tabcolsep{3pt}
			\renewcommand\arraystretch{1.1}
                \begin{tabular}{cc|cccc}
				\hline\thickhline
				Aggregation  & Distance & Attribute & Layout  & Non-spatial & Complex \\
				\hline\hline
                \multicolumn{6}{l}{\textit{Baseline (DreamLLM~\cite{dong2023dreamllm})}} \\
                - & \multicolumn{1}{c|}{-}  & 22.94 & 23.74 & 28.76 & 23.01 \\
                \hline
                \multicolumn{6}{l}{\textit{Supervised Fine-tuning (SFT)}} \\
                \hline
                - & \multicolumn{1}{c|}{Eucl.}  & 12.25 & 0.75  & 16.41 & 11.71 \\
                - & \multicolumn{1}{c|}{Cos}  & 6.95  & 0.29  & 16.78 & 11.48 \\
                AvgPool & \multicolumn{1}{c|}{Eucl.}  & 23.31 & 23.89 & 28.76 & 23.22 \\
                AvgPool & \multicolumn{1}{c|}{Cos}  & 23.12 & 24.20  & 28.79 & 23.29 \\
                \hline
                \multicolumn{6}{l}{\textit{Continuous Kernel-based Direct Preference Optimization}} \\
                \hline
                - & \multicolumn{1}{c|}{Eucl.}  & 23.65 & 24.34 & 28.83 & 23.08 \\
                - & \multicolumn{1}{c|}{Cos}  & 23.97 & 24.11 & 28.77 & 23.21 \\
                MaxPool & \multicolumn{1}{c|}{Eucl.}  & 23.79 & 24.04 & 28.86 & 23.92 \\
                MaxPool & \multicolumn{1}{c|}{Cos}  & 29.18 & 25.01 & 18.72 & 12.27 \\
                AvgPool & \multicolumn{1}{c|}{Eucl.}  & 26.75 & 24.70  & \textbf{28.94} & 25.12 \\
                \rowcolor{gray_bg}
                AvgPool & \multicolumn{1}{c|}{Cos}  & \textbf{34.85} & \textbf{25.51} & 28.82 &\textbf{ 25.31} \\
                \hline
			\end{tabular}
		}
            }
	}
\vspace{-3ex}
\end{table}

\mysubsubsec{Kernel-based Continuous DPO}. 
In Sec.~\ref{sec:kc_dpo}, we introduce the KC-DPO to fine-tune LMMs with continuous representations. The implementation of kernel functions can be divided into \textit{Aggregation} and \textit{Distance}. To assess the impacts of different kernels, we conduct extensive comparison experiments, as shown in Tab.~\ref{tab:abla_kernel}. We observe SFT slightly improves the alignment performance, while DPO yields more substantial gains across all metrics. These results show that kernel functions are crucial to KC-DPO, and an optimal choice could greatly enhance the efficiency of preference optimization in continuous feature space. Overall, AvgPool + Cos demonstrates the superior performance improvement.

\begin{figure}[h]
\centering
 \subfloat[$\beta$ in Discrete DPO]{
\label{fig:beta_seed}
\includegraphics[scale=0.19]{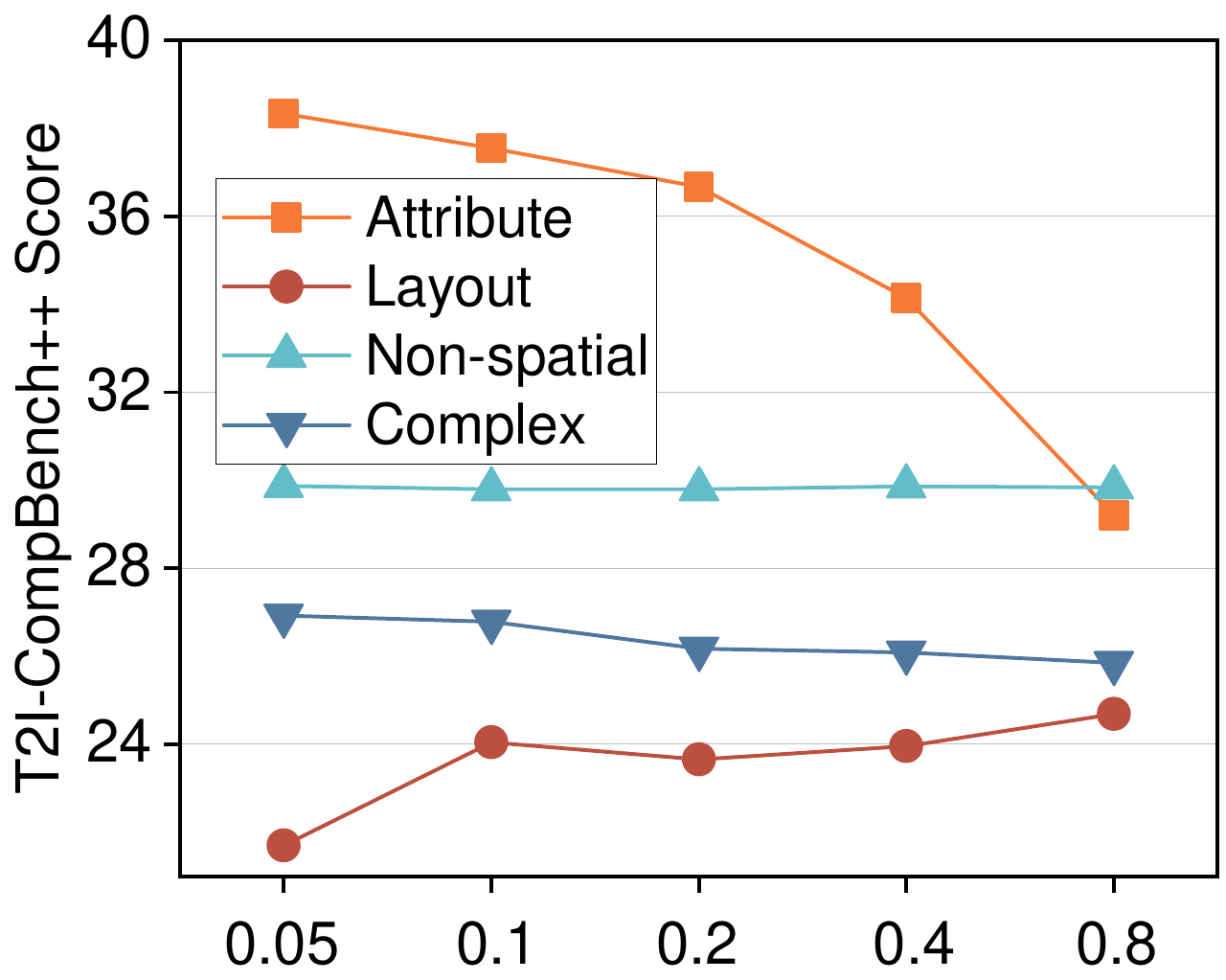}}\hspace{-0.15cm}
\subfloat[$\gamma$ in Continuous KC-DPO]{
\label{fig:gamma_dreamllm}
\includegraphics[scale=0.179]{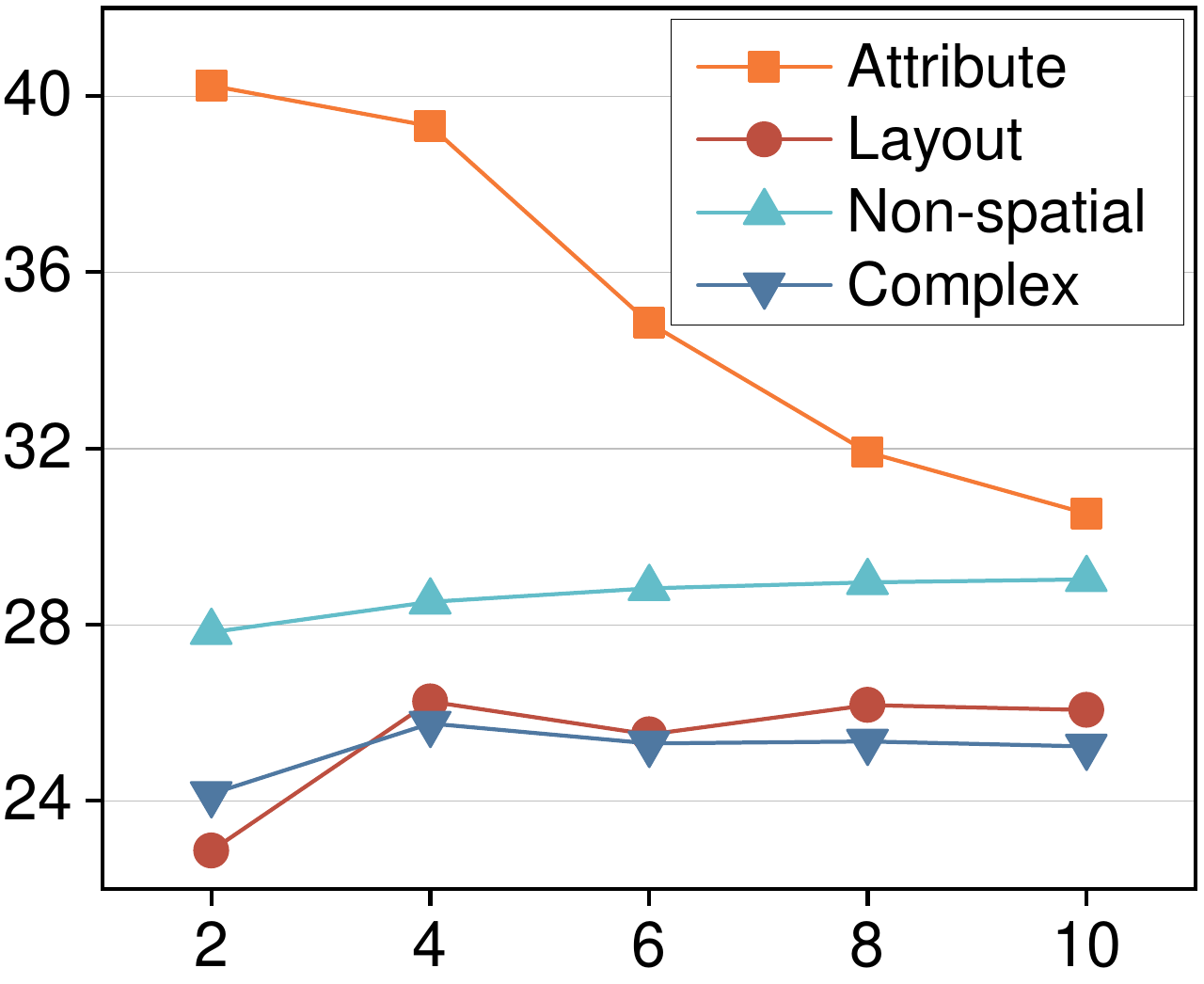}}\vspace{-0.1cm} 
\vspace{-1ex}
\caption{Hyperparameter sensitivity on four general categories of T2I-CompBench++, examining (a) $\beta$ in discrete DPO for SEED-LLaMA, and (b) $\gamma$ in continuous KC-DPO for DreamLLM.}
\label{fig:beta_and_gamma}
\vspace{-2ex}
\end{figure}

\begin{figure}[t]
        \centering
	\includegraphics[width=0.43\textwidth]{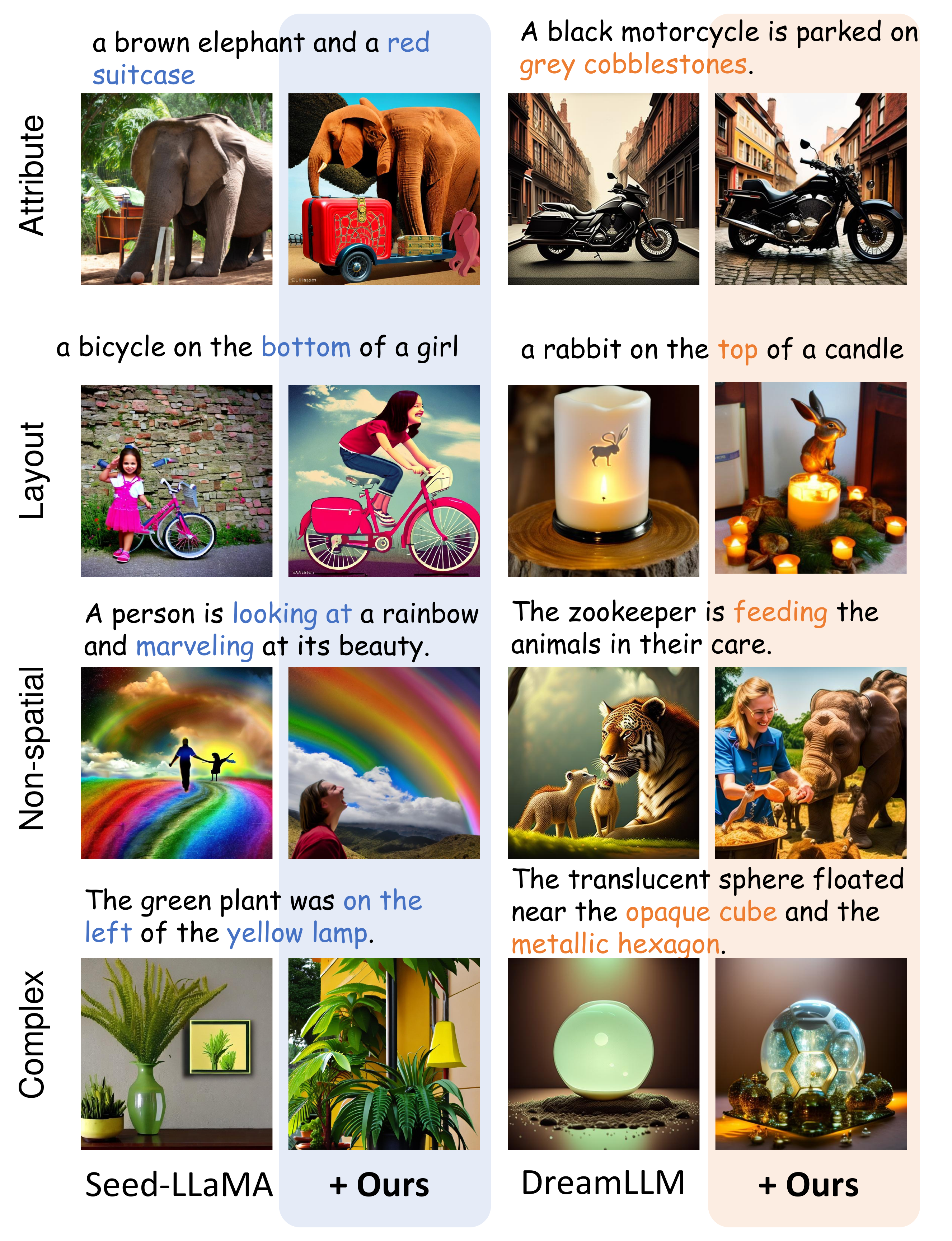}
	\vspace{-2ex}
	\caption{Qualitative results from SEED-LLaMA, DreamLLM, and the proposed SILMM method, on T2I-CompBench++. }
	\label{fig:qual}
	\vspace{-3ex}
\end{figure}
\subsection{Qualitative Results}
To illustrate the improvements achieved by SILMM, Fig.~\ref{fig:qual} presents examples generated by SEED-LLaMA, DreamLLM, and SILMM, in Fig.~\ref{fig:qual} using prompts from T2I-CompBench++. These results showcase the effectiveness of SILMM across extensive compositional scenarios. 

\section{Conclusion}\label{sec:conclu}
In this work, we present a self-improvement approach named SILMM to enhance text-image alignment within LMMs, introducing an iterative model-agnostic framework comprising five stages to enable high-quality self-feedback and alignment learning. For continuous LMMs, we propose a dropout-based strategy to diversify image representations and a continuous DPO method, KC-DPO, for optimizing LMMs with preference representation pairs. Extensive experiments validate the effectiveness and superiority of our SILMM framework.
{
    \small
    \bibliographystyle{ieeenat_fullname}
    \bibliography{main}
}
\clearpage
\maketitlesupplementary

\section{Details of Compositional Prompt Generation}\label{sec:comp_prompt}
For attribute and layout prompt generation, we first leverage the world knowledge of LMMs to generate common objects spanning various categories, including animals, plants, fruits, household items, clothing, vehicles, food, musical instruments, and electronic devices. 
Attributes such as color, shape, texture, and 2D/3D spatial relations are also incorporated. 
Using predefined templates, we systematically combine objects with attributes, numeracy, and spatial relations to construct compositional prompts. The templates are detailed below:

\mysubsubsec{Attribute.} 
\begin{itemize}
    \item \textit{A \{adj\} \{noun\}}
    \item \textit{A \{adj1\} \{noun1\} and a \{adj2\} \{noun2\}}
\end{itemize} 

\mysubsubsec{Layout}

\begin{itemize}
    \item \textit{A \{noun1\} \{spatial\_2d/spatial\_3d\} a \{noun2\}}
    \item \textit{\{quantity\} \{object\_singular/object\_plural\}}
    \item \textit{\{quantity\} \{object\_singular/object\_plural\} and \{quantity\} \{object\_singular/object\_plural\}} 
\end{itemize}

For non-spatial and complex relations, we adopt in-context learning to generate diverse prompts based on LMMs:

\begin{tcolorbox}[colback=gray!5,colframe=gray!90,title=Instruction for Non-spatial Prompt Generation]
\textbf{System Prompt}

{\small
You are an assistant dedicated to generating natural prompts that contain subjects and objects by using non-spatial relationship words such as wear, watch, speak, hold, have, run, look at, talk to, jump, play, walk with, stand on, and sit on.
}

\vspace{0.7em} 

\textbf{User Prompt}

{\small
Input: Generate a prompt that contains subjects and objects by using non-spatial relationship words.

Output: Two friends are watching a movie together on a large TV screen.

\vspace{0.7em}

Input: Generate a prompt that contains subjects and objects by using non-spatial relationship words.

Output: Two athletes are running along the beach as the sun sets behind them.

\vspace{0.7em}

Input: Generate a prompt that contains subjects and objects by using non-spatial relationship words.

Output: 
}

\end{tcolorbox}

\begin{tcolorbox}[colback=gray!5,colframe=gray!90,title=Instruction for Complex Prompt Generation]
\textbf{System Prompt}

{\small
You are an assistant dedicated to generating natural compositional phrases or prompts, containing multiple objects (number $\geq$ 2) with one or several adjectives from color, shape, and texture descriptions and spatial (left/right/top/bottom/next to/near/on side of) or non-spatial relationships. 
}

\vspace{0.7em} 

\textbf{User Prompt}

{\small
Input: Please generate a compositional phrase or sentence containing multiple objects with one or several adjectives and relationships. 

Output: The fluffy white cat sat next to the black leather couch.

\vspace{0.7em}

Input: Please generate a compositional phrase or sentence containing multiple objects with one or several adjectives and relationships. 

Output: The sleek black phone rested beside the textured brown leather wallet.

\vspace{0.7em}

Input: Please generate a compositional phrase or sentence containing multiple objects with one or several adjectives and relationships. 

Output: The red spherical balloon floated above the striped rectangular kite and the green triangular flag.

\vspace{0.7em}

Input: Please generate a compositional phrase or sentence containing multiple objects with one or several adjectives and relationships. 

Output: The golden, sunlit leaves floated softly above the jagged, rust-colored rocks, their delicate, lacy shapes casting playful shadows on the uneven ground.

\vspace{0.7em}

Input: Please generate a compositional phrase or sentence containing multiple objects with one or several adjectives and relationships. 

Output: 
}

\end{tcolorbox}

\section{Details of Self-Questioning Prompt}\label{sec:self_q_prompt}
We follow a divide-and-conquer strategy, where the LMM first extracts the atomic concepts from the given prompt. These atomic concepts are then transformed into simple yes-or-no questions. The specific instructions are shown in the following: 

\begin{tcolorbox}[colback=gray!5,colframe=gray!90,title={Instruction for Self-Questioning on Attribute (Color, Shape, and Texture) Prompt}]
\textbf{System Prompt}

{\small
You are an assistant dedicated to transforming a sentence into several questions. You should first divide it into simple concepts and relations, and then provide the corresponding questions. Avoid using pronouns, such as he, she, it, and they.
}

\vspace{0.7em} 

\textbf{User Prompt}

{\small
Input: A white harp and a rust soup.

Output: Concepts and relations: a white harp, a rust soup;
Questions: Is there a white harp? Is there a rust soup?

\vspace{0.7em}

Input: A quarter circle lily and a hexagon mirror.

Output: Concepts and relations: a quarter circle lily, a hexagon mirror;
Questions: Is there a quarter-circle lily? Is there a hexagon mirror?

\vspace{0.7em}

Input: Shiny mop and metal key holder.

Output: Concepts and relations: a shiny mop, a metal key holder;
Questions: Is there a shiny mop? Is there a metal key holder?

\vspace{0.7em}

Input: \{prompt\}

Output: 
}

\end{tcolorbox}

\begin{tcolorbox}[colback=gray!5,colframe=gray!90,title={Instruction for Self-Questioning on Layout (Spatial, 3D-Spatial, and Numeracy) Prompt}]
\textbf{System Prompt}

{\small
You are an assistant dedicated to transforming a sentence into several questions. You should first divide it into simple concepts and relations, and then provide the corresponding questions. Avoid using pronouns, such as he, she, it, and they.
}

\vspace{0.7em} 

\textbf{User Prompt}

{\small
Input: A pancake on the left of a pasta.

Output: Concepts and relations:  a pancake, a pasta, a pancake is on the left of a pasta;
Questions: Is there a pancake? Is there a pasta? Is a pancake on the left of a pasta?

\vspace{0.7em}

Input: A lamp behind a screwdriver.

Output: Concepts and relations: a lamp, a screwdriver, a lamp is behind a screwdriver;
Questions: Is there a lamp? Is there a screwdriver? Is a lamp behind a screwdriver?

\vspace{0.7em}

Input: Three light bulbs and eight pumpkins.

Output: Concepts and relations: three light bulbs, eight pumpkins; 
Questions: Are there three light bulbs? Are there eight pumpkins? 

\vspace{0.7em}

Input: \{prompt\}

Output: 
}

\end{tcolorbox}

\begin{tcolorbox}[colback=gray!5,colframe=gray!90,title={Instruction for Self-Questioning on Non-Spatial and Complex Prompt}]
\textbf{System Prompt}

{\small
You are an assistant dedicated to transforming a sentence into several questions. You should first divide it into simple concepts and relations, and then provide the corresponding questions. Avoid using pronouns, such as he, she, it, and they.
}

\vspace{0.7em} 

\textbf{User Prompt}

{\small
Input: A chef is holding a knife and preparing a dish on the stove. 

Output: Concepts and relations: a chef, a knife, a dish, the stove, a chef is holding a
knife, a chef is preparing a dish;
Questions: Is there a chef? Is there a knife? Is there a dish? Is there a stove?
Is a chef holding a knife? Is a chef preparing a dish?

\vspace{0.7em}

Input: The green teapot is located near the round oak table. 

Output: Concepts and relations: a green teapot, a round oak table, the green teapot is near
the round oak table, the round oak table is near the green teapot;
Questions: Is there a green teapot? Is there a round oak table? Is the green teapot near the
round oak table? Is the round oak table near the green teapot?

\vspace{0.7em}

Input: The chunky wooden lamp casts a warm glow on the tattered blue curtains.

Output: Concepts and relations: a chunky wooden lamp, a warm glow, tattered blue curtains,
a chunky wooden lamp casts a warm glow, the warm glow is on the tattered blue curtains;
Questions: Is there a chunky wooden lamp? Is there a warm glow? Are there tattered blue
curtains? Is a chunky wooden lamp casting a warm glow? Is the warm glow on the tattered
blue curtains?

\vspace{0.7em}

Input: The vibrant orange tomato sat atop the crisp green leaf and the juicy red watermelon.

Output: Concepts and relations: a vibrant orange tomato, a crisp green leaf, a juicy red
watermelon, a vibrant orange tomato is atop a crisp green leaf, a vibrant orange tomato is
atop a juicy red watermelon;
Questions: Is there a vibrant orange tomato? Is there a crisp green leaf? Is there a juicy red
watermelon? Is the vibrant orange tomato atop the crisp green leaf? Is the vibrant orange
tomato atop the juicy red watermelon?

\vspace{0.7em}

Input: \{prompt\}

Output: 
}

\end{tcolorbox}

\section{Derivation of KC-DPO}\label{sec:kc_dpo_deriv}
\subsection{Preliminary}
\mysubsubsec{Reinforcement Learning from Feedback with Reward Model}. 
With collected preference pairs $\mathcal{D} = \{(x^i, y_w^i, y_l^i)\}_{i=1}^N$ from human feedback~\cite{ouyang2022training} or AI feedback~\cite{lee2024rlaif, yu2024rlaif}, a reward model $r_\phi (x, y)$ is trained to maximize the likelihood~\cite{rafailov2024direct}: 
\begin{equation}
    p_\phi(y_w \succ y_l) = \frac{\exp(r_\phi(x, y_w))}{\exp(r_\phi(x, y_w) + \exp(r_\phi(x, y_l))}, 
\end{equation}
where $y_w$ and $y_l$ denote the preferred and dispreferred responses. The likelihood maximization objective can be implemented by minimizing the following loss for binary classification~\cite{rafailov2024direct}: 
\begin{equation}
    \mathcal{L}_R = - \mathbb{E}_{(x, y_w, y_l) \sim \mathcal{D}} [\log \sigma (r_\phi (x, y_w) - r(x, y_l))], 
\end{equation}
where $\sigma$ denotes a sigmoid function. After the training phase, the reward model could provide a reward value as feedback for any prompt-response pair $(x, y)$ on the fly. 

Based on the feedback from the reward model, a language model $\pi_\theta$ can be optimized via RL fine-tuning~\cite{rafailov2024direct, jaques2017sequence, jaques2020human}, which is formulated as: 
\begin{equation}
    \max_{\pi_\theta} \mathbb{E}_{x \sim \mathcal{D}, y \sim \pi_\theta(y|x)} [r_\phi (x, y)] - \beta \rm KL(\pi_\theta (y | x) || \pi_{ref} (y | x)), 
\end{equation}
where $\beta$ controls the strength of following the distribution of the reference model and avoids potential risks of model degradation. $\rm{KL(\cdot || \cdot)}$ refers to Kullback–Leibler divergence. The language model can not be directly optimized by gradient descent using this objective because of the discreteness of language. Existing work~\cite{ziegler2019fine, stiennon2020learning, ouyang2022training,bai2022training} adopts RL, specifically the PPO~\cite{schulman2017proximal} algorithm, to maximize the reward function: 
\begin{equation}
    r(x, y) = r_\phi (x, y) - \beta (\log \pi_\theta (y | x) - \log \pi_{ref} (y | x)). 
\end{equation}

\mysubsubsec{Direct Preference Optimization}. 
Though the above two-stage learning strategy has achieved remarkable progress~\cite{ouyang2022training, touvron2023llama2}, it requires training a reward model and the final performance highly depends on it. To alleviate such dependency, DPO~\cite{rafailov2024direct} was proposed by deriving a closed form of the preference optimization process, which avoids training a reward model. The DPO method uses an alternative parameterization to learn an implicit reward and the loss is written as: 
\begin{multline}\label{eqn:app_dpo}
\small 
    \mathcal{L}_{\text{DPO}} = -\mathbb{E}_{(x, z_w, z_l) \sim \mathcal{D}} \\
    \left[\log \sigma \left(\beta \log \frac{\pi_\theta (z_w | x)}{\piref (z_w | x)} - \beta \log \frac{\pi_\theta (z_l | x)}{\piref (z_l | x)}\right)\right]. 
\end{multline}

\subsection{Kernel-based Continuous DPO}
The DPO objective is proposed for optimizing language models which represent language as discrete tokens, and model token distributions as categorical distributions. Such discreteness and categorical distribution modeling make it simple to calculate the likelihood $\pi(y|x)$ in DPO. As discussed in Sec.~\ref{sec:kc_dpo}, however, it is intractable to calculate the likelihood $\pi(\bm{H}|x)$ for continuous LMMs where $\bm{H}$ denotes a continuous feature. 

To model the distribution of the intermediate continuous feature, we first decomposite the log-likelihood per time step and make the Gaussian assumption as, 
\begin{equation}\label{eqn:app_gaussian}
\small 
\begin{aligned}[t]
    &~~~~~\log \pi(\bm{H} \mid x) \\
    &= \sum_{i=1}^{L} \log \pi(\bm{h}_i \mid \bm{H}_{<i}, x) \\
    &= \sum_{i=1}^{L} \log \frac{\exp\left(-\frac{1}{2}(\bm{h}_i - \bm{\mu}_i)^\top \bm{\Sigma}_i^{-1} (\bm{h}_i - \bm{\mu}_i)\right)}{\sqrt{(2\pi)^D |\bm{\Sigma}_i|}} \\
    &= \sum_{i=1}^{L} \left[-\frac{1}{2} (\bm{h}_i - \bm{\mu}_i)^\top \bm{\Sigma}_i^{-1} (\bm{h}_i - \bm{\mu}_i)\right] - \sum_{i=1}^{L} \log \sqrt{(2\pi)^D |\bm{\Sigma}_i|}, 
\end{aligned}
\end{equation}
where $L$ denotes the sequence length of the continuous feature\footnote{To preserve visual details, continuous LMMs~\cite{dong2023dreamllm, sun2023emu, ge2024seed} often represent a continuous feature with a sequence of feature vectors. For example, $L = 64$ in DreamLLM~\cite{dong2023dreamllm}. } and $D$ refers to the feature dimension. 

We assume that the Gaussian distribution is isotropic and all dimensions share the same variance value $\bar{\sigma}$, \ie, $\bm{\Sigma}_i \approx \rm{diag} (\sigma_1, ..., \sigma_D)$ and $\sigma_1 = ... = \sigma_D = \bar{\sigma}$, attaining: 
\begin{equation}\label{eqn:app_appro}
\small 
\begin{aligned}[t]
    &~~~~~\log \pi(\bm{H} \mid x) \\
    &= \sum_{i=1}^{L} \left[-\frac{1}{2} (\bm{h}_i - \bm{\mu}_i)^\top \bm{\Sigma}_i^{-1} (\bm{h}_i - \bm{\mu}_i)\right] - \sum_{i=1}^{L} \log \sqrt{(2\pi)^D |\bm{\Sigma}_i|} \\
    &\approx -\frac{1}{2\bar{\sigma}}\sum_{i=1}^{L} \left[(\bm{h}_i - \bm{\mu}_i)^\top (\bm{h}_i - \bm{\mu}_i) \right] - \frac{D}{2} \sum_{i=1}^{L} \log 2\pi\bar{\sigma} \\
    &= -\frac{1}{2\bar{\sigma}}\sum_{i=1}^{L} \Vert\bm{h}_i - \bm{\mu}_i\Vert^2_2 - C.  \\
\end{aligned}
\end{equation}
The above simplification reformulates the likelihood into an L2-norm expression due to the Gaussian assumption. 

Next, with the simplified likelihood of continuous features, we induce the continuous DPO by substituting Eqn.~(\ref{eqn:app_appro}) into Eqn.~(\ref{eqn:app_dpo}):

\begin{equation}\label{eqn:app_cont_dpo}
\small 
\begin{aligned}[t]
    &~~~~~\mathcal{L}_{\text{DPO}} \\
    &= -\mathbb{E}_{(x, z_w, z_l) \sim \mathcal{D}} \left[\log \sigma \left(\beta \log \frac{\pi_\theta (z_w | x)}{\piref (z_w | x)} - \beta \log \frac{\pi_\theta (z_l | x)}{\piref (z_l | x)}\right)\right] \\
    &\approx -\mathbb{E}_{(x, z_w, z_l) \sim \mathcal{D}} \Bigg[\log \sigma\left( -\frac{\bar{\sigma}\beta}{2} \sum_{i=1}^{L} \Vert\bm{h}_i^w - \bm{\mu}_i\Vert^2_2 - \beta C \right. \\
    &\hspace{2cm} \left. + \frac{\beta}{2\bar{\sigma}} \sum_{i=1}^{L} \Vert\bm{h}_i^w - \bm{\mu}_i^{ref}\Vert^2_2 + \beta C \right. \\
    &\hspace{2cm} \left. - \frac{\beta}{2\bar{\sigma}} \sum_{i=1}^{L} \Vert\bm{h}_i^l - \bm{\mu}_i\Vert^2_2 - \beta C \right. \\
    &\hspace{2cm} \left. + \frac{\beta}{2\bar{\sigma}} \sum_{i=1}^{L} \Vert\bm{h}_i^l - \bm{\mu}_i^{ref}\Vert^2_2 + \beta C \right) \Bigg] \\
    &= -\mathbb{E}_{(x, z_w, z_l) \sim \mathcal{D}} \Bigg[\log \sigma\Big(\frac{\beta}{2\bar{\sigma}} \sum_{i=1}^{L}(-\Vert\bm{h}_i^w - \bm{\mu}_i\Vert^2_2 +\Vert\bm{h}_i^w - \bm{\mu}_i^{ref}\Vert^2_2) \\
    &\hspace{2cm}  - \Vert\bm{h}_i^l - \bm{\mu}_i\Vert^2_2 + \Vert\bm{h}_i^l - \bm{\mu}_i^{ref}\Vert^2_2\Big) \Bigg] \\
    &\approx  -\mathbb{E}_{(x, z_w, z_l) \sim \mathcal{D}} \Bigg[\log \sigma\Big(\frac{\beta}{2\bar{\sigma}} (
    -\Vert \bm{H} - \bm{H}_w \Vert_F^2 \nonumber 
    + \Vert \bm{H}_r - \bm{H}_w \Vert_F^2) \\
    &\hspace{2cm} + \Vert \bm{H} - \bm{H}_l \Vert_F^2 
    - \Vert \bm{H}_r - \bm{H}_l \Vert_F^2 \Big) \Bigg], 
\end{aligned}
\end{equation}
where we make $\bm{\mu}_i \approx \bm{h}_i$ and $\bm{\mu}_i^{ref} \approx \bm{h}_i^{ref}$, \ie, we approximate the mean vector with the online output of the policy network and the reference network. 

Finally, we introduce the kernel function theory and obtain a generalized form of the continuous DPO: 
\begingroup
\small 
\setlength{\jot}{-5pt} 
\begin{align}\label{eqn:app_kc_dpo}
    \mathcal{L}_{\text{KC-DPO}} &= -\mathbb{E}_{(x, \bm{H}_w, \bm{H}_l) \sim \mathcal{D}} 
    \Bigg[\log\sigma \bigg( \gamma 
    (-k(\bm{H}, \bm{H}_w) \nonumber \\
    & \quad + k(\bm{H}_r, \bm{H}_w) 
     + k(\bm{H}, \bm{H}_l)  
     - k(\bm{H}_r, \bm{H}_l) ) \bigg) \Bigg], 
\end{align}
\endgroup
where $\gamma = \frac{\beta}{\bar{\sigma}^2}$ is a hyperparameter that controls the balance between the reference model and preference optimization. A higher value of $\gamma$ encourages the optimized policy model to adhere to the reference model more closely. 
$k(\cdot, \cdot)$ represents a generalized distance measurement function, and the objective formulated in Eqn.~(\ref{eqn:app_kc_dpo}) is named as Kernel-based Continuous DPO (KC-DPO).

\section{Implementation Details}\label{sec:app_dpo_train}
We employ Low-Rank Adaptation (LoRA)~\cite{hu2021lora} for efficient optimization of SEED-LLaMA and DreamLLM, using the same LoRA settings for both models, with a LoRA rank and hyperparameter $\alpha$ of 32. 
For SEED-LLaMA, the LLM backbone of DreamLLM is optimized for 1k steps, with a learning rate of $5 \times 10^{-5}$, 100 warm-up steps, and a cosine learning rate scheduler.
The batch size is set to 32 with a gradient accumulation step of 4. 
The $\beta$ hyperparameter in DPO (Eqn.~(\ref{eqn:dpo})) is set to 0.2. 

For DreamLLM, training is conducted for 2k steps with a learning rate of $8 \times 10^{-6}$, 200 warm-up steps, and the same cosine learning rate scheduler. The batch size and gradient accumulation step remain at 32 and 4, respectively. The adherence degree $\gamma$ in KC-DPO (Eqn.~($\ref{eqn:kc_dpo}$)) is set to 3.0.

\section{DPO Training Data}\label{sec:dpo_data}
In each iteration, SEED-LLaMA and DreamLLM are instructed to generate 16k prompts encompassing a wide spectrum of compositional scenarios, as detailed in Step 1 of Sec.~\ref{sec:silmm}. 
For discrete optimization of SEED-LLaMA, we generate 10 images per prompt, selecting the top-ranked and last-ranked representations\textemdash scored via VQA-based self-feedback\textemdash as the chosen and rejected pairwise training samples, respectively. 

For continuous optimization of DreamLLM, to improve tuning stability, we generate 30 images per prompt and select the top 10 and last 10 representations as chosen and rejected samples. These are combined to produce 100 pairs per prompt. 

\section{Additional Experimental Results}\label{sec:add_exp}
\subsection{Additional Quantitative Results}
\mysubsubsec{Performance Improvement over Iterations}. 
We show the performance improvement of the proposed SILMM method over three iterations, on detailed categories of T2I-CompBench++~\cite{huang2023t2i}, DPG-Bench~\cite{hu2024ella}, and TIFA~\cite{hu2023tifa}, as shown in Tab.~\ref{tab:app_perf_t2icompbench}, Tab.~\ref{tab:app_perf_dpgbench}, Tab.~\ref{tab:app_perf_tifa}, respectively. These results demonstrate that the proposed method yields improvements across most categories as the iteration progresses. However, due to limitations in multiple capabilities—such as prompt generation, decompositional question generation, VQA-based self-feedback, and basic visual generation—the rate of improvement slows and may eventually reach a saturation point. 

\begin{figure}[t]
        \centering
	\includegraphics[width=0.37\textwidth]{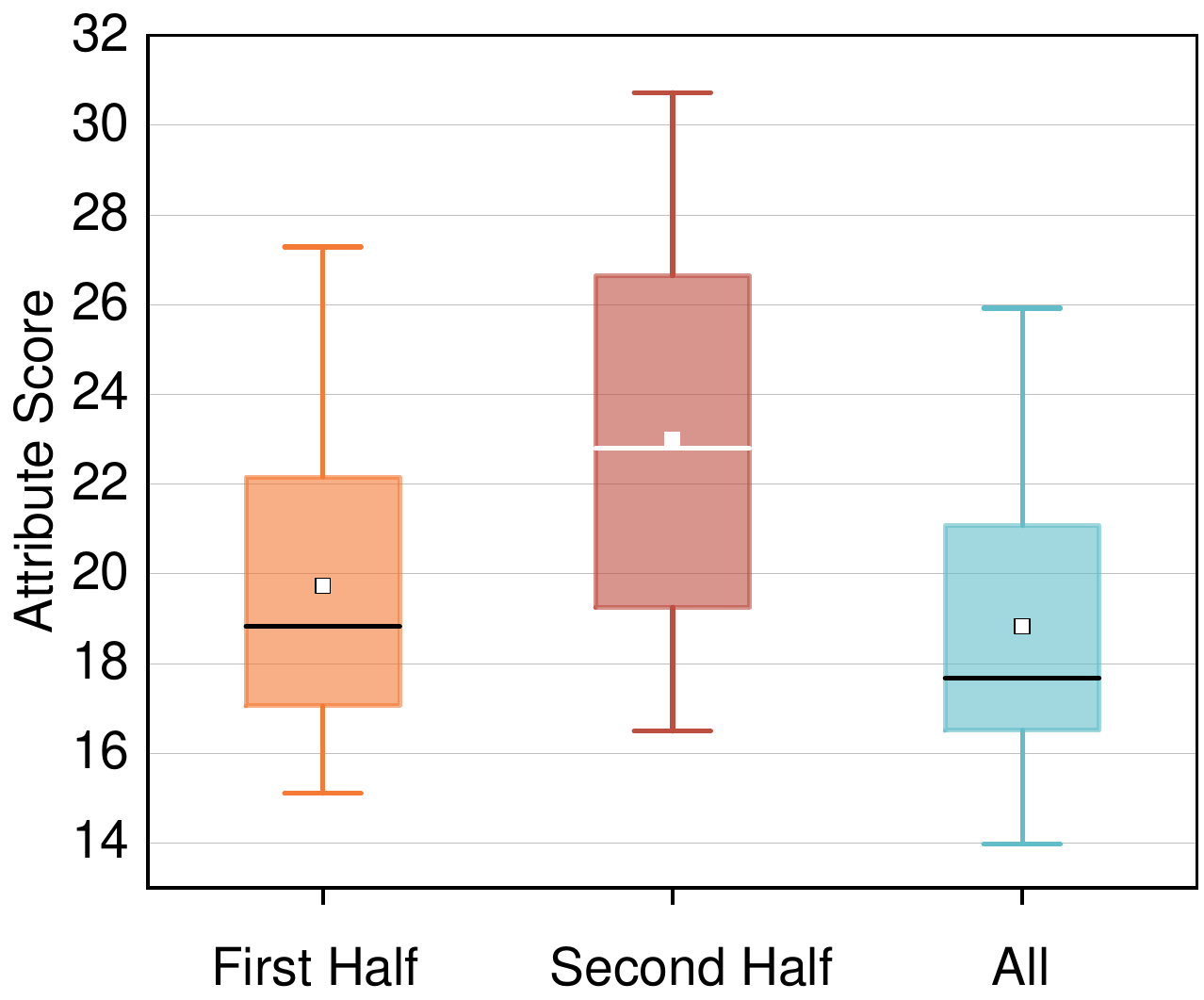}
	\vspace{-2ex}
	\caption{Comparison of perturbing different layers of LMMs for diverse continuous representation generation on Color, Shape, and Texture categories of T2I-CompBench++. }
	\label{fig:app_div_layers_att}
	\vspace{-3ex}
\end{figure}

\begin{figure}[t]
        \centering
	\includegraphics[width=0.37\textwidth]{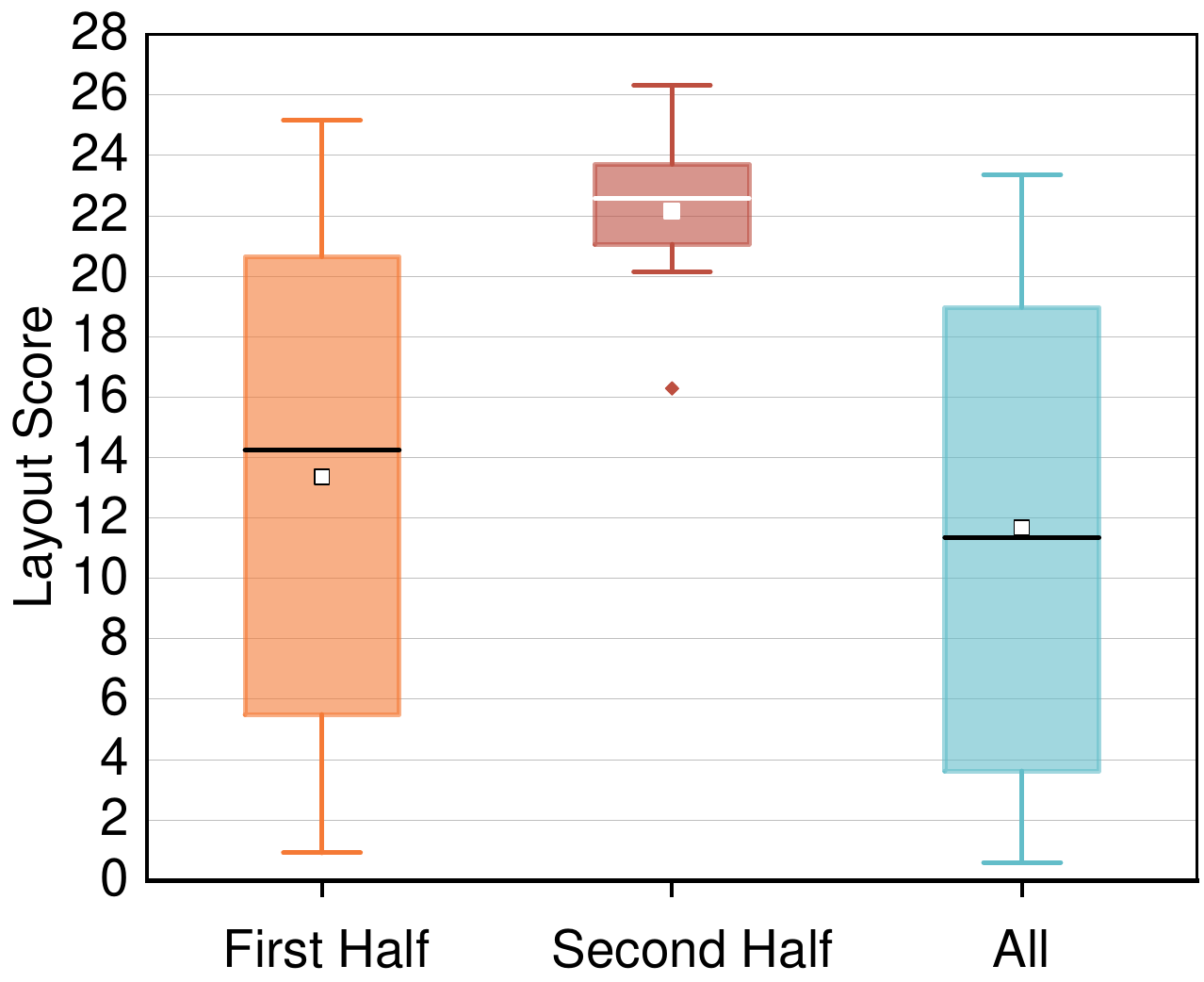}
	\vspace{-2ex}
	\caption{Comparison of perturbing different layers of LMMs for diverse continuous representation generation on Spatial, 3D Spatial, and Numeracy categories of T2I-CompBench++. }
	\label{fig:app_div_layers_layout}
	\vspace{-3ex}
\end{figure}

\begin{figure}[t]
        \centering
	\includegraphics[width=0.37\textwidth]{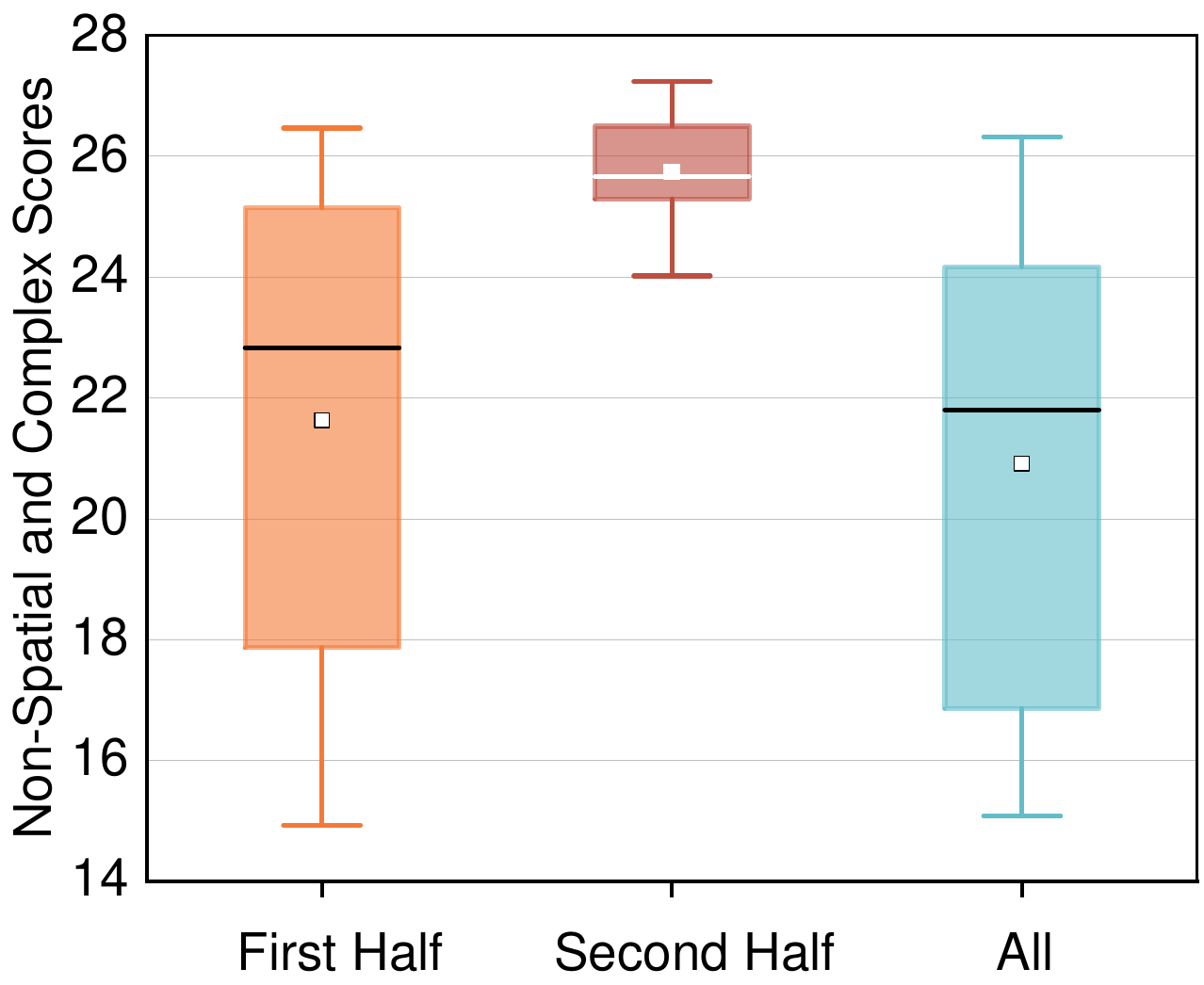}
	\vspace{-2ex}
	\caption{Comparison of perturbing different layers of LMMs for diverse continuous representation generation on Non-spatial and Complex categories of T2I-CompBench++. }
	\label{fig:app_div_layers_noncomp}
	\vspace{-3ex}
\end{figure}

\mysubsubsec{In-depth Analysis of DropDiv}. 
Fig.~\ref{fig:app_div_layers_att}, Fig.~\ref{fig:app_div_layers_layout}, and Fig.~\ref{fig:app_div_layers_noncomp} present comparisons of three settings of DropDiv for generating diverse continuous representations, with alignment scores evaluated on the validation set of T2I-CompBench++. 
``First Half'', ``Second Half'', and ``All'' represent adding and performing dropout operations in the first (bottom) layers, the last (top) layers, and all layers of DreamLLM. 
Each prompt in the dataset is used to generate ten distinct representations and corresponding images using DreamLLM. The figure is divided into three sections: (a) Color, Shape, and Texture, (b) Spatial, 3D Spatial, and Numeracy, and (c) Non-spatial and Complex. 

\mysubsubsec{In-depth Analysis of Negative Sampling}. 
In Tab.~\ref{tab:app_neg_samp}, we compare different negative sampling ranges on 8 categories of T2I-CompBench++. The results show that different negative sampling ranges may have different influences for different categories. For instance, soft sampling is beneficial to the attribute categories while may not be the best choice for numeracy and non-spatial categories. 

\begin{table*}[t]
	\centering
	\setlength{\abovecaptionskip}{0.15cm}
	\caption{Performance improvement of the proposed SILMM method over three iterations (Iter.) for compositional text-to-image generation on the 8 categories of the T2I-CompBench++~\cite{huang2023t2i} benchmark. Alignment scores are calculated using expert understanding models (\eg, VQA or object detection models) recommended by T2I-CompBench++~\cite{huang2023t2i}. 
 }
	\label{tab:app_perf_t2icompbench}
	{\hspace{-1ex}
             \setlength{\tabcolsep}{1mm}{
		\resizebox{0.83\textwidth}{!}
		{
			\setlength\tabcolsep{8pt}
			\renewcommand\arraystretch{1.1}
                \begin{tabular}{l|ccc|ccc|c|c}
				\hline\thickhline
				\multicolumn{1}{c|}{\multirow{2}{*}{Method}} & \multicolumn{3}{c|}{Attribute}  & \multicolumn{3}{c|}{Layout} & \multicolumn{1}{c|}{\multirow{2}{*}{Non-spatial}} & \multicolumn{1}{c}{\multirow{2}{*}{Complex}}  \\ 
				\cline{2-7}
				 & Color & Shape & \multicolumn{1}{l|}{Texture} & Spatial & 3D Spatial & \multicolumn{1}{l|}{Numeracy} &   &    \\
				\hline\hline
				\multicolumn{1}{l|}{SEED-LLaMA~\cite{ge2024making}} & 17.87 & 19.43 & 20.31 & 5.72  & 21.72 & 33.43 & 28.86 & 21.46 \\
                \multicolumn{1}{l|}{~ + SILMM (Iter. 1)} & 37.41 & 33.12 & 39.46 & 9.16  & \textbf{26.07} & 35.75 & 29.80  & 26.17 \\
                \multicolumn{1}{l|}{~~~~ + SILMM (Iter. 2)} & 39.81 & \textbf{37.62} & 38.00 & 8.60 & 25.42 & \textbf{38.59} & 29.62 & 27.14 \\
                \rowcolor{gray_bg}
                \multicolumn{1}{l|}{~~~~~~~ + SILMM (Iter. 3)} & \textbf{41.91} & 36.27 & \textbf{40.63} & \textbf{11.90}  & 25.74 & 37.70  & \textbf{29.82} & \textbf{28.28} \\
                \hline
                \multicolumn{1}{l|}{DreamLLM~\cite{dong2023dreamllm}} & 21.04 & 21.86 & 25.91 & 6.13  & 25.62 & 39.46 & 28.76 & 23.01 \\
                \multicolumn{1}{l|}{~ + SILMM (Iter. 1)} & 32.47 & 32.25 & 39.84 & 8.87  & 27.60  & 40.07 & 28.82 & 25.31 \\
                \multicolumn{1}{l|}{~~~~ + SILMM (Iter. 2)} & \textbf{36.39} & 35.82 & 47.28 & 12.13 & 27.76 & 41.44 & 28.94 & \textbf{26.87} \\
                \rowcolor{gray_bg}
                \multicolumn{1}{l|}{~~~~~~~ + SILMM (Iter. 3)} & 35.61 & \textbf{36.83} & \textbf{47.39} & \textbf{12.70}  & \textbf{28.58} & \textbf{41.61} & \textbf{29.00} & 26.43 \\
				\hline
			\end{tabular}
		}
            }
	}
\vspace{-2ex}
\end{table*}

\begin{table*}[t]
	\centering
	\setlength{\abovecaptionskip}{0.15cm}
	\caption{Performance improvement of the proposed SILMM method over three iterations (Iter.) for complex text-to-image generation on the 5 categories of the DPG-Bench~\cite{hu2024ella} benchmark. Alignment scores are calculated using expert understanding models (\eg, VQA or object detection models) recommended by DPG-Bench. 
 }
	\label{tab:app_perf_dpgbench}
	{\hspace{-1ex}
             \setlength{\tabcolsep}{1mm}{
		\resizebox{0.77\textwidth}{!}
		{
			\setlength\tabcolsep{15pt}
			\renewcommand\arraystretch{1.1}
                \begin{tabular}{l|ccccc|c}
				\hline\thickhline
				\multicolumn{1}{c|}{\multirow{1}{*}{Method}} & Color & Shape & \multicolumn{1}{l}{Texture} & Spatial & 3D Spatial  & All \\
				\hline\hline
				\multicolumn{1}{l|}{SEED-LLaMA~\cite{ge2024making}} & 65.59 & 55.87 & 62.00    & 62.77 & 59.46 & 47.12 \\
                \multicolumn{1}{l|}{~ + SILMM (Iter. 1)} & 69.73 & 70.33 & 69.40  & 73.27 & 68.65 & 57.07 \\
                \multicolumn{1}{l|}{~~~~ + SILMM (Iter. 2)} & 73.41 & 69.04 & \textbf{71.00} & 74.47 & \textbf{69.18} & 56.94 \\
                \rowcolor{gray_bg}
                \multicolumn{1}{l|}{~~~~~~~ + SILMM (Iter. 3)} & \textbf{73.55} & \textbf{70.48} & 68.50  & \textbf{74.79} & 68.64 & \textbf{57.31} \\
                \hline
                \multicolumn{1}{l|}{DreamLLM~\cite{dong2023dreamllm}} & 74.47 & 65.86 & 63.80  & 74.24 & 46.00    & 53.93 \\
                \multicolumn{1}{l|}{~ + SILMM (Iter. 1)} & 74.47 & 73.31 & 67.00    & 80.39 & 52.80  & 60.95 \\
                \multicolumn{1}{l|}{~~~~ + SILMM (Iter. 2)} & 75.38 & \textbf{76.61} & \textbf{69.20} & \textbf{84.41} & \textbf{62.40} & \textbf{64.47} \\
                \rowcolor{gray_bg}
                \multicolumn{1}{l|}{~~~~~~~ + SILMM (Iter. 3)} & \textbf{76.29} & 75.91 & \textbf{69.20} & \textbf{84.41} & 60.00    & 64.22 \\
				\hline
			\end{tabular}
		}
            }
	}
\vspace{-2ex}
\end{table*}

\begin{table*}[t]
	\centering
	\setlength{\abovecaptionskip}{0.15cm}
	\caption{Performance improvement of the proposed SILMM method over three iterations (Iter.) for compositional text-to-image generation on the 12 categories of the TIFA~\cite{hu2023tifa} benchmark. Alignment scores are calculated using expert understanding models (\eg, VQA or object detection models) recommended by TIFA. 
 }
	\label{tab:app_perf_tifa}
	{\hspace{-1ex}
             \setlength{\tabcolsep}{1mm}{
		\resizebox{0.95\textwidth}{!}
		{
			\setlength\tabcolsep{3.5pt}
			\renewcommand\arraystretch{1.1}
                \begin{tabular}{l|cccccccccccc|c}
				\hline\thickhline
				\multicolumn{1}{c|}{\multirow{1}{*}{Method}} & Animal & Object & Location & Activity & Color & Spatial & Attribute & Food  & Counting & Material & Other & Shape & ALL \\
				\hline\hline
				\multicolumn{1}{l|}{SEED-LLaMA~\cite{ge2024making}} & 69.35 & 63.14 & 72.55 & 65.73 & 60.59 & 66.75 & 71.9  & 60.37 & 61.66 & 68.42 & 52.74 & 43.48 & 66.74 \\
                \multicolumn{1}{l|}{~ + SILMM (Iter. 1)} & 76.52 & 71.67 & 75.27 & \textbf{74.5} & 74.7  & 72.36 & 74.52 & 66.85 & 65.82 & 75.16 & 60.7  & 52.17 & 73.82 \\
                \multicolumn{1}{l|}{~~~~ + SILMM (Iter. 2)} & 76.75 & \textbf{72.65} & \textbf{76.41} & 73.87 & \textbf{78.03} & \textbf{71.35} & \textbf{75.46} & 67.18 & \textbf{65.92} & \textbf{81.82} & \textbf{64.18} & 56.52 & \textbf{74.47} \\
                \rowcolor{gray_bg}
                \multicolumn{1}{l|}{~~~~~~~ + SILMM (Iter. 3)} & \textbf{76.98} & 72.1  & 74.89 & 73.38 & 77.91 & 71.13 & 73.08 & \textbf{70.36} & 63.29 & 78.95 & \textbf{64.18} & \textbf{62.32} & 73.74 \\
                \hline
                \multicolumn{1}{l|}{DreamLLM~\cite{dong2023dreamllm}} & 75.44 & 67.7  & 75.6  & 64.64 & 63.57 & 67.24 & 70.43 & 70.69 & 61.05 & 75.6  & 55.22 & 56.52 & 69.91 \\
                \multicolumn{1}{l|}{~ + SILMM (Iter. 1)} & 78.81  & 71.67  & 79.35  & 72.26  & 63.74  & 71.48  & 72.70  & 73.55  & 61.97  & 75.60  & 61.19  & 63.77  & 73.37  \\
                \multicolumn{1}{l|}{~~~~ + SILMM (Iter. 2)} & 80.06  & \textbf{74.28 } & \textbf{79.57 } & \textbf{76.18 } & 63.74  & \textbf{75.54 } & \textbf{74.40 } & 76.51  & \textbf{66.73 } & \textbf{77.99 } & \textbf{68.66 } & 60.87  & \textbf{75.59 } \\
                \rowcolor{gray_bg}
                \multicolumn{1}{l|}{~~~~~~~ + SILMM (Iter. 3)} & \textbf{80.29 } & 73.85  & 79.35  & 75.34  & \textbf{63.80 } & 74.53  & 74.05  & \textbf{77.06 } & 65.72  & 77.51  & 67.66  & \textbf{65.22 } & 75.38  \\
				\hline
			\end{tabular}
		}
            }
	}
\vspace{-2ex}
\end{table*}

\begin{table*}[t]
	\centering
	\setlength{\abovecaptionskip}{0.15cm}
	\caption{Influence of negative sampling for KC-DPO on the 8 categories of the T2I-CompBench++~\cite{huang2023t2i} benchmark.. ``14 - 24'' means the rejected data points are sampled from rank-14 to rank-24 which is a hard range, while ``20 - 30'' refers to the last 10 samples which is the softest range. We generate 30 images per prompt. 
 }
	\label{tab:app_neg_samp}
	{\hspace{-1ex}
             \setlength{\tabcolsep}{1mm}{
		\resizebox{0.85\textwidth}{!}
		{
			\setlength\tabcolsep{8pt}
			\renewcommand\arraystretch{1.1}
                \begin{tabular}{l|ccc|ccc|c|c}
				\hline\thickhline
				\multicolumn{1}{c|}{\multirow{2}{*}{Negative Range}} & \multicolumn{3}{c|}{Attribute}  & \multicolumn{3}{c|}{Layout} & \multicolumn{1}{c|}{\multirow{2}{*}{Non-spatial}} & \multicolumn{1}{c}{\multirow{2}{*}{Complex}}  \\ 
				\cline{2-7}
				 & Color & Shape & \multicolumn{1}{l|}{Texture} & Spatial & 3D Spatial & \multicolumn{1}{l|}{Numeracy} &   &    \\
				\hline\hline
				\multicolumn{1}{c|}{14 - 24} & 23.58 & 26.03 & 31.02 & 7.65  & 27.44 & \textbf{41.47} & \textbf{29.08} & 24.83 \\
                \multicolumn{1}{c|}{16 - 26} & 25.57 & 26.13 & 32.70  & 8.28  & 27.28 & 40.27 & 29.06 & 24.85 \\
                \multicolumn{1}{c|}{18 - 28} & 27.06 & 27.44 & 34.72 & \textbf{8.92}  & 26.84 & 40.56 & 28.86 & \textbf{25.57} \\
                \rowcolor{gray_bg}
                \multicolumn{1}{c|}{20 - 30} & \textbf{32.47} & \textbf{32.25} & \textbf{39.84} & 8.87  & \textbf{27.60}  & 40.07 & 28.82 & 25.31 \\
				\hline
			\end{tabular}
		}
            }
	}
\vspace{-2ex}
\end{table*}

\subsection{Additional Qualitative Results}
\begin{figure*}[t]
        \centering
        \vspace{-4ex}
	\includegraphics[width=0.99\textwidth]{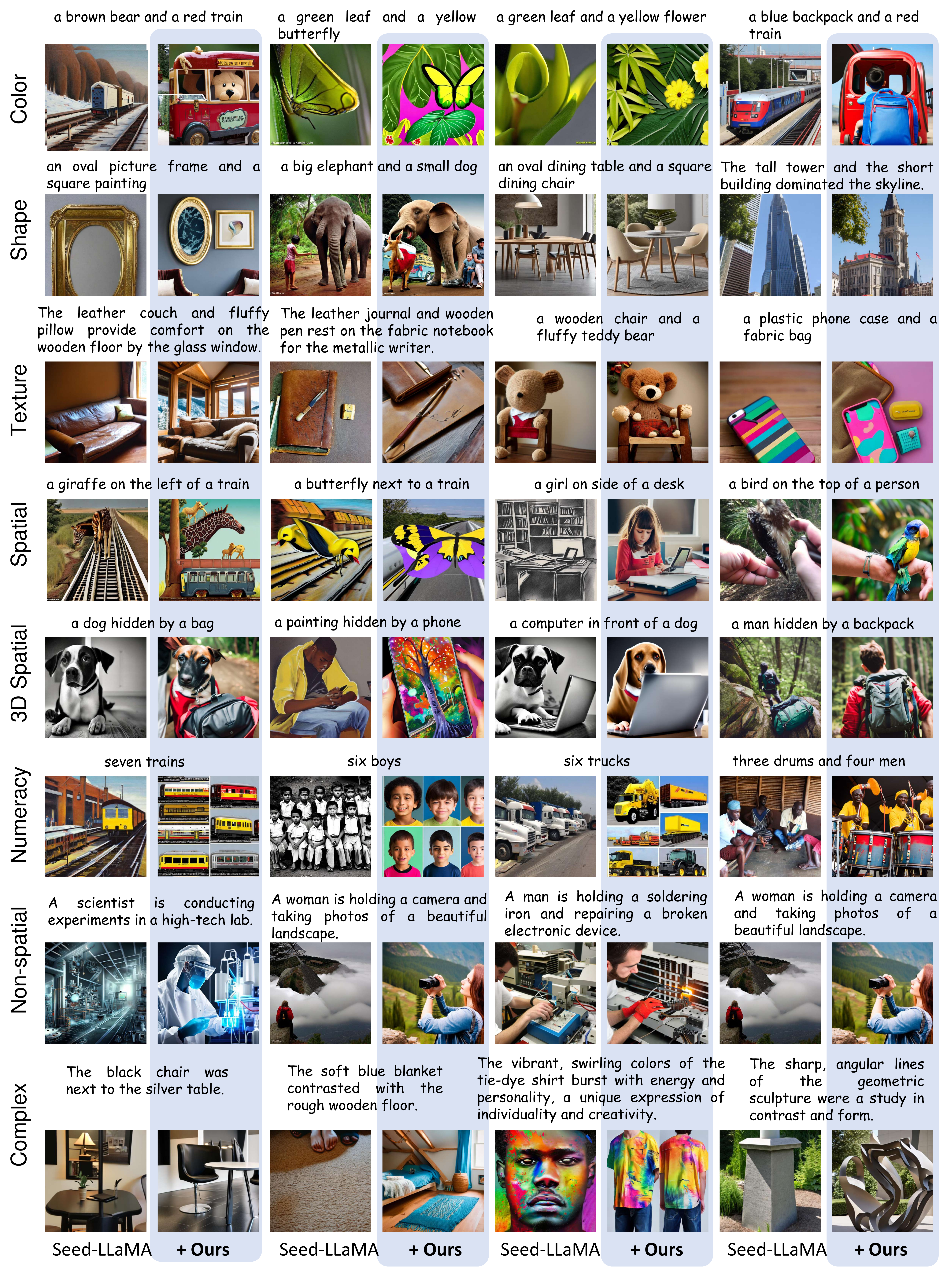}
	\vspace{-3ex}
	\caption{Qualitative results of SEED-LLaMA and the proposed SILMM method on the T2I-CompBench++~\cite{huang2023t2i} benchmark. }
	\label{fig:app_qua_seedllama_t2icompbench}
	\vspace{-3ex}
\end{figure*}

\begin{figure*}[t]
        \centering
	\includegraphics[width=0.99\textwidth]{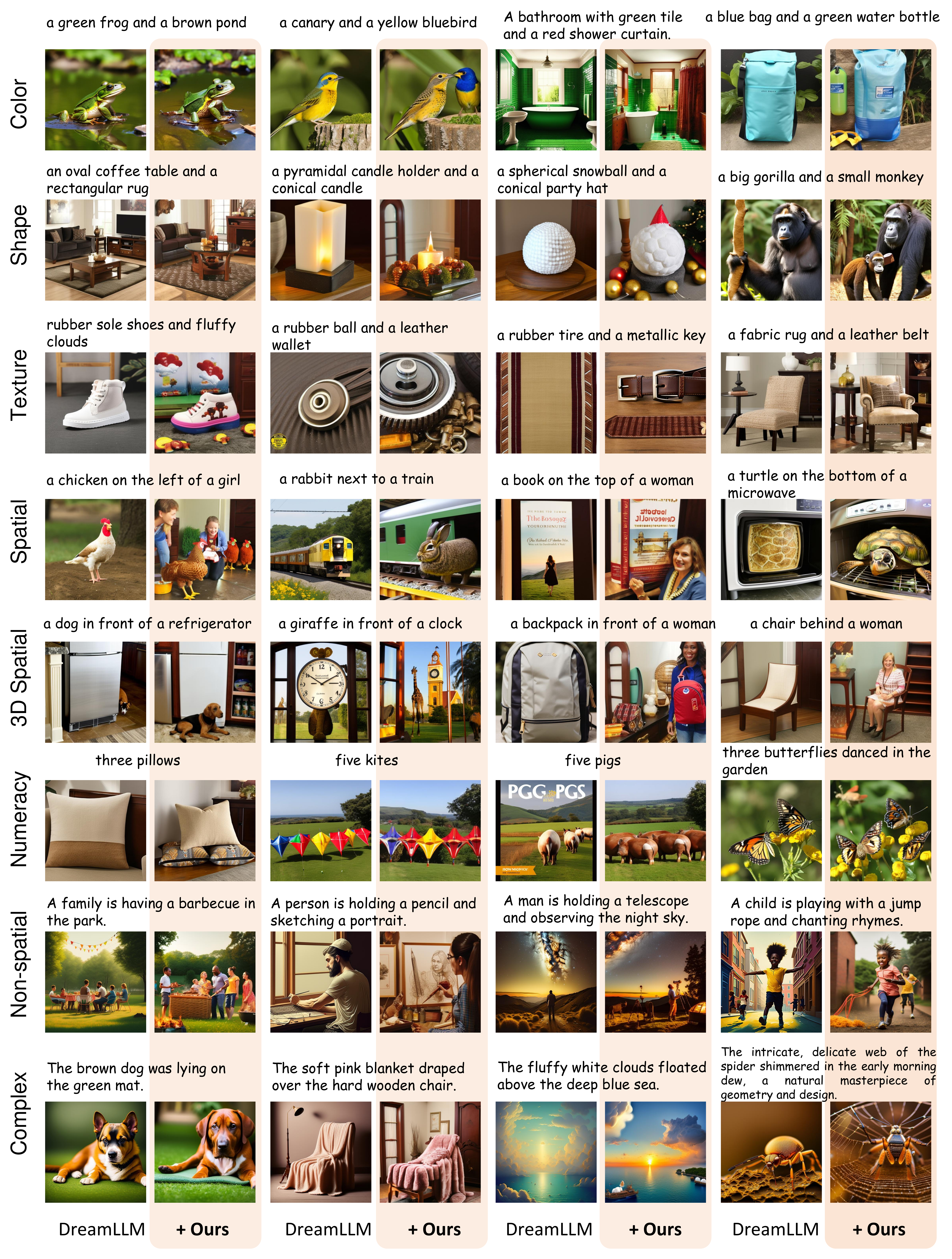}
	\vspace{-2ex}
	\caption{Qualitative results of DreamLLM and the proposed SILMM method on the T2I-CompBench++~\cite{huang2023t2i} benchmark. }
	\label{fig:app_qua_dreamllm_t2icompbench}
	\vspace{-3ex}
\end{figure*}

\begin{figure*}[t]
        \centering
        \vspace{-4ex}
	\includegraphics[width=0.99\textwidth]{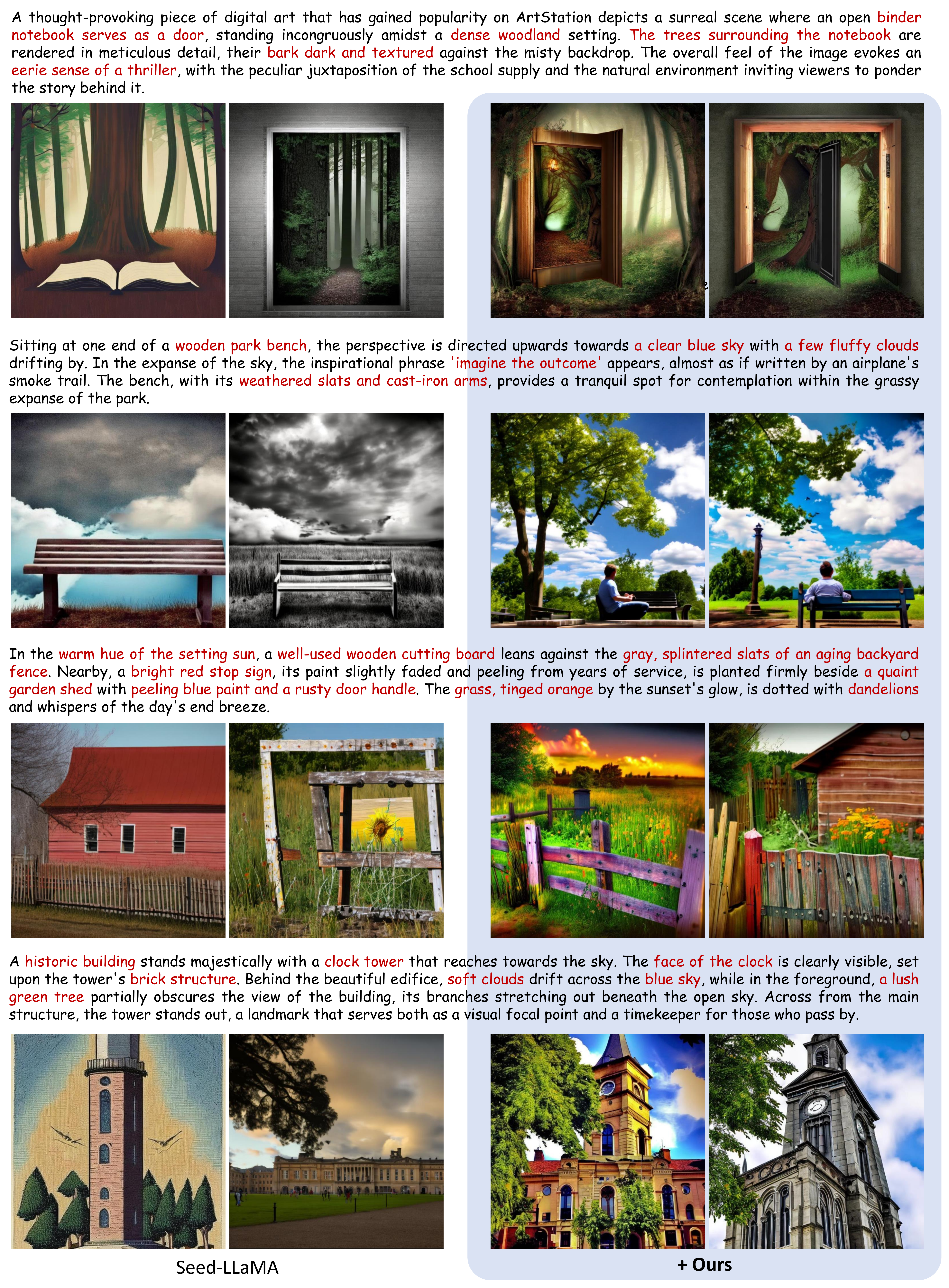}
	\vspace{-3ex}
	\caption{Qualitative results of SEED-LLaMA and the proposed SILMM method on the DPG-Bench~\cite{hu2024ella} benchmark. }
	\label{fig:app_qua_seedllama_dpgbench}
	\vspace{-3ex}
\end{figure*}

\begin{figure*}[t]
        \centering
        \vspace{-4ex}
	\includegraphics[width=0.99\textwidth]{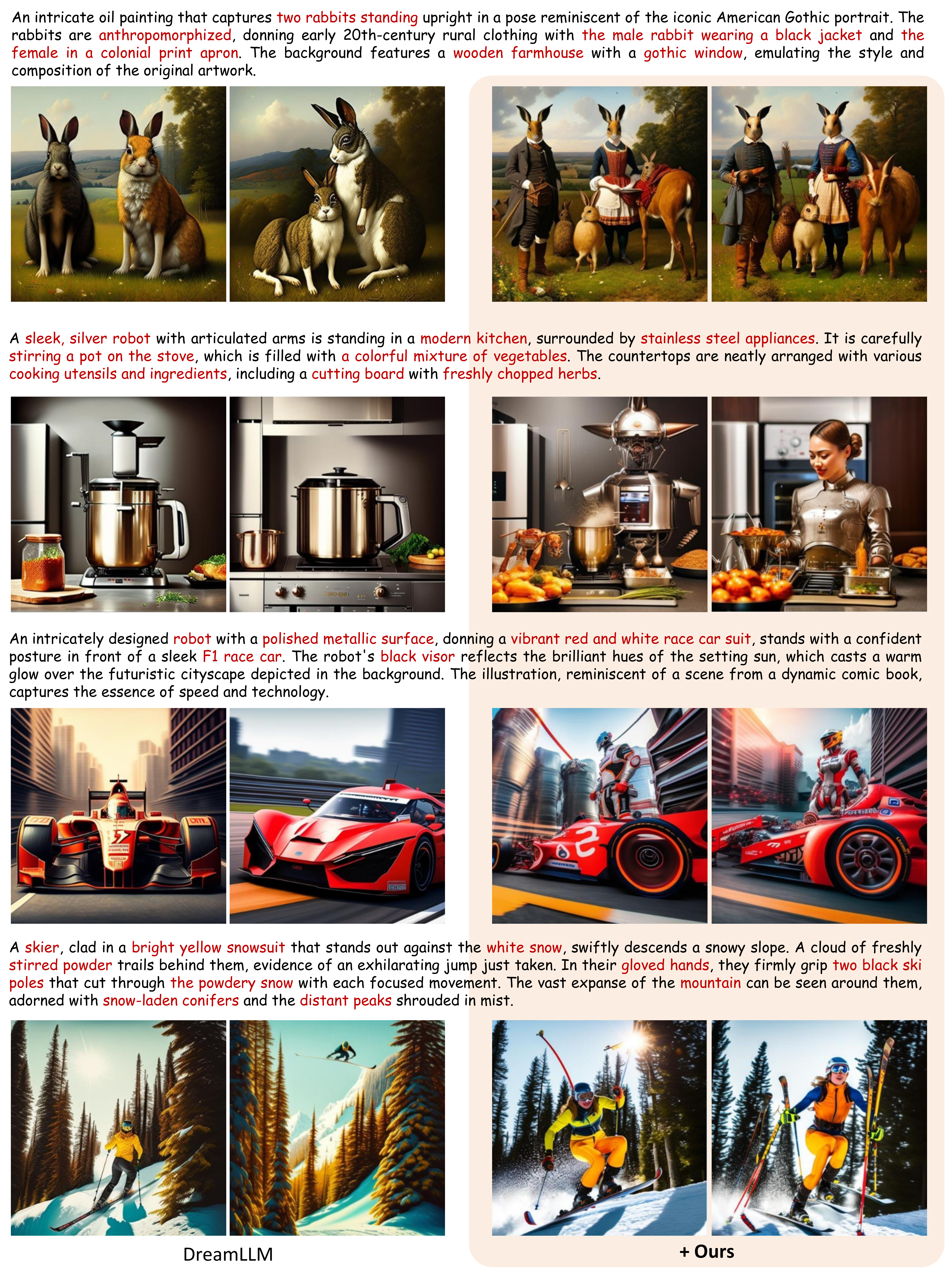}
	\vspace{-3ex}
	\caption{Qualitative results of DreamLLM and the proposed SILMM method on the DPG-Bench~\cite{hu2024ella} benchmark. }
	\label{fig:app_qua_dreamllm_dpgbench}
	\vspace{-3ex}
\end{figure*}

There has been a surge of research interests in tackling the challenging cross-modal misalignment~\cite{feng2023training, wen2021comprehensive, qin20243d, chen2023multimodal, ADDRL2024MM} problem in the multimodal learning community.
To intuitively understand the improvement of SILMM on text-image alignment in compositional or complex scenarios, we list some images generated by SEED-LLaMA and SILMM on T2I-CompBench++~\cite{huang2023t2i} in Fig.~\ref{fig:app_qua_seedllama_t2icompbench}, and images generated by DreamLLM and SILMM in Fig.~\ref{fig:app_qua_dreamllm_t2icompbench}. Besides, we also show examples on the recent benchmark DPG-Bench~\cite{hu2024ella} which contains more challenging long and complex prompts in Fig.~\ref{fig:app_qua_seedllama_dpgbench} and Fig.~\ref{fig:app_qua_dreamllm_dpgbench}. 

As shown in these visual examples, SILMM consistently outperforms the base models, \ie, SEED-LLaMA and DreamLLM in terms of text-image alignment, especially in more compositional and complex scenarios. In the images generated by SEED-LLaMA and DreamLLM, we observe noticeable misalignments and inaccuracies when handling intricate relationships between objects and scene details. In contrast, SILMM is able to produce more coherent and contextually accurate images, demonstrating its effectiveness across different compositional scenarios, especially long-form and highly descriptive ones. 

\section{Future Work}
In future work, we aim to enhance the efficiency of LMMs for image synthesis through strategies such as efficient tuning~\cite{guo2024learning, luo2024cheap} and accelerated inference~\cite{shang2024llava}. Additionally, we plan to investigate the interplay between intrinsic understanding and generative capabilities in LMMs, aiming to foster their mutual enhancement.

\end{document}